%% file: main_arxiv.tex
\documentclass{article}
\usepackage{amsmath,epsfig,bm, amssymb,subfigure,graphicx,algorithm,algorithmic,color,natbib}
\input{mathdef}

\usepackage{url,multirow}

\advance\oddsidemargin-1.0in
\date{\today}
\title{Post Selection Inference with Kernels} 

\author{
Makoto Yamada$^1$, Yuta Umezu$^2$, Kenji Fukumizu$^3$, Ichiro Takeuchi$^2$\\
$^1$Kyoto University, Kyoto, Japan \\
$^2$Nagoya Institute of Technology, Nagoya, Japan \\
$^3$Institute of Statistical Mathematics, Tokyo, Japan \\
\texttt{makoto.m.yamada@ieee.org}, \texttt{\{umezu.yuta,takeuchi.ichiro\}@nitech.ac.jp}\\
\texttt{fukumizu@ism.ac.jp}
}

\begin{document}
\maketitle

\begin{abstract}
 We propose a novel kernel based post selection inference (PSI) algorithm, which can not only handle non-linearity in data but also structured output such as multi-dimensional and multi-label outputs. Specifically, we develop a PSI algorithm for independence measures,  and propose the Hilbert-Schmidt Independence Criterion (HSIC) based PSI algorithm (\texttt{hsicInf}). The novelty of the proposed algorithm is that it can handle non-linearity and/or structured data through kernels. Namely, the proposed algorithm can be used for wider range of applications including nonlinear multi-class classification and multi-variate regressions, while existing PSI algorithms cannot handle them. Through synthetic experiments, we show that the proposed approach can find a  set of \emph{statistically significant} features for both regression and classification problems. Moreover, we apply the \texttt{hsicInf} algorithm to a real-world data, and show that \texttt{hsicInf} can successfully identify important features. 
\end{abstract}

\section{Introduction} 
Finding a set of features in high-dimensional data is an important problem with many real-world problems such as biomarker discovery \citep{xing2001feature}, document categorization \citep{CIKM:Forman:2008}, and prosthesis control \citep{IEEEBIO:Shenoy+etal:2008}, to name a few. In particular, finding a  set of \emph{statistically significant} features is crucial for scientific discovery, and linear methods including the least absolute shrinkage and selection operator (LASSO) \citep{JRSSB:Tibshirani:1996} are extensively used. However, LASSO focuses on finding a set of \emph{linearly} related features. Thus, if an input and output pair has \emph{non-linear} relationship, it is hard to find a set of important features. 

To select non-linearly related features, a feature screening approach, which is based on ranking features with respect to the association score between each feature and its output,  is widely used \citep{fan2008sure}. Typically, the correlation coefficient (linear) and the mutual information (non-linear) are used as an association measure \citep{book:Cover+Thomas:2006}. Recently, the kernel based independence measure such as the Hilbert-Schmidt Independence Criterion (HSIC) \citep{ALT:Gretton+etal:2005} and its normalized variant (NOCCO) were proposed and have started being used as a surrogate of the mutual information \citep{song2012feature,balasubramanian2013ultrahigh,NIPS:Fukumizu+etal:2008}.  The key advantage of the kernel based approaches is that it can deal with non-linearity and/or non-structured data including multi-variate and graph data  through kernels. Besides association approaches, sparse regularization based approaches including the sparse additive model (SpAM) \citep{ravikumar2009sparse}  and HSIC LASSO \citep{yamada2014high} are widely used in feature selection communities.  These kernel based approaches are promising, however, it is not clear whether the selected features are \emph{statistically significant}. Since the \emph{selection event} needs to be taken into account for statistical inference, a naive two-step approach, which first selects features and then test the selected features without any adjustment, does not control the desired false positive rates.

The problem of testing the significance of the selected features is known as \emph{selective inference} \citep{taylor2015statistical,hastie2015statistical}. The data splitting approach is a typical selective inference algorithm. The key idea of the splitting approach is to divide a training dataset into two disjoint sets, and then, one of the sets is used for feature selection and the other is used for statistical inference. Since the selection event and the statistical inference are independent, we can successfully identify a set of statistically important features. However, since we divide a dataset into two disjoint sets and used them for feature selection and statistical inference, respectively, the detection power can be degraded.

Recently, a novel selective inference approach called the post selection inference (PSI) has been proposed  \citep{hastie2015statistical,lockhart2014significance,lee2016exact}. PSI algorithms tend to have higher detection power than data splitting approaches. However, only \emph{linear} approaches which are built upon LASSO or other similar linear feature selection approaches are available so far. Since real-world datasets tend to have non-linear relationship,  the existing linear approaches may fail to find a set of important features; this is a critical problem in practice. Moreover, existing PSI approaches are only applicable to uni-variate output. Thus, the applications of existing PSI methods is limited. 

In this paper, we propose a kernel based PSI method \texttt{hsicInf}, which can find \emph{statistically significant} features from non-linear and/or structured output data such as multi-dimensional output. Specifically, we develop a PSI algorithm for independence measures, and propose the HSIC based PSI algorithm. A clear advantage of \texttt{hsicInf} over existing approaches is that it can easily handle non-linearity and structured data through kernels. Namely, it can be used for wider range of applications including multi-class classification and multi-variate regression. Through synthetic and real-world experiments, we show that the proposed approach can find a set of \emph{statistically significant} features for both regression and classification problems.

\noindent {\bf Contribution:}
\begin{itemize}
\item We propose a kernel based PSI algorithm that  can  handle non-linearity, multi-label, multi-class, and structured output. To the best of our knowledge, this paper presents the first kernel based PSI algorithm.
\item We empirically show that the proposed algorithm can  successfully identify a set of non-linearly related features. 
\end{itemize}

\section{Proposed method}
In this section, we propose a PSI method with kernels. More specifically, we develop a new PSI framework based on an independence measure called the Hilbert-Schmidt Independence Criterion (HSIC) \citep{ALT:Gretton+etal:2005,zhang2016large}. 

\subsection{Problem Formulation}
Let us denote an input vector by $\boldx = [x^{(1)}, \dots, x^{(d)}]^\top \in \mathbbR^d$ and  the corresponding target vector $\boldy \in \mathbbR^{d_y}$. i.i.d. samples $\{(\boldx_i, \boldy_i)\}_{i = 1}^{\ntr}$ have been drawn from a joint probability density $p(\boldx, \boldy)$. 
The final goal of this paper is to first screen $k < d$ features of input vector $\boldx$ and then test whether the selected features are of \emph{statistically significant} association to its output $\boldy$. 

\subsection{Marginal screening and post-selection inference with independence measure}


In this paper, we employ an estimate of the independence measure $\widehat{I}(X_m,Y)$, which measures the discrepancy from the independence  between the $m$-th random variable $X_m$ and its output variable $Y$, where the vector of independence measures denoted by $\boldz = [\widehat{I}(X_1, Y), \ldots, \widehat{I}(X_d, Y)]^\top$ follows a multi-variate normal distribution with $\boldmu \in \mathbbR^d$ and $\boldSigma \in \mathbbR^{d \times d}$: 
\[
\boldz \sim N(\boldmu, \boldSigma).
\]
Then, we exploit the normality of $\boldz$ and combine it with the post selection inference framework recently developed by \citep{lee2016exact} (see Theorem~\ref{theo1}). 



\begin{theo} 
\label{theo1}
\citep{lee2016exact}. Consider a stochastic data generating process $\boldcalZ \sim N(\boldmu, \boldSigma)$. If a feature selection event is characterized by $\boldA \boldcalZ \leq \boldb$ for a matrix $\boldA$ and a vector $\boldb$ that do not depend on $\boldcalZ$, then, for any fixed vector $\boldeta \in \mathbbR^{d}$,
\begin{align*}
F^{[V^-(\boldA,\boldb),V^+(\boldA,\boldb) ]}_{\boldeta^\top \boldmu, \boldeta^\top \boldSigma \boldeta} (\boldeta^\top \boldcalZ) ~~|~~ \boldA \boldcalZ \leq \boldb \sim \textnormal{Unif}(0,1),
\end{align*}
where $F^{[v,w]}_{t,u}(\cdot)$ is the cumulative distribution function of the uni-variate truncated normal distribution with the mean $t$, the variance u, and the lower and the upper truncation points $v$ and $w$, respectively. Furthermore, using $\boldc := \frac{\boldSigma \boldeta}{\boldeta^\top \boldSigma \boldeta}$, the lower and the upper truncation points are given as
\begin{align}
\label{eq:lower_psi}
V^-(\boldA,\boldb) &:= \max_{j:(\boldA\boldc)_j < 0}\left\{ \frac{b_j - (\boldA\boldz)_j}{(\boldA\boldc)_j}\right\} + \boldeta^\top \boldz, \\
V^+(\boldA,\boldb) &:= \min_{j:(\boldA\boldc)_j > 0}\left\{ \frac{b_j - (\boldA\boldz)_j}{(\boldA\boldc)_j}\right\} + \boldeta^\top \boldz.
\label{eq:upper_psi}
\end{align}
\proofend
\end{theo}

In order to develop post selection inference method for the independence measure $\boldz$, we confirm that the problem of selecting top $k$ features in the decreasing order of $\widehat{I}(X_{\cdot}, Y)$ can be represented as a linear selection event in the form of $\boldA \boldz \le \boldb$ in Theorem~\ref{theo1}. 

We denote the index set of the selected $k$ features by $\calS$, and that of the unselected $\bar{k} = d - k$ features by $\bar{\calS}$. The fact that $k$ features in $\calS$ are selected and $\bar{k}$ features in $\bar{\calS}$ are not selected is rephrased by
\begin{align}
\label{eq:selection_event}
\widehat{I}(X_m,Y) \geq \widehat{I}(X_\ell,Y), ~~\text{for all}~(m,\ell) \in \calS \times \bar{\calS}.
\end{align}
Here, we have in total $k\bar{k}$ constraints written as the linear inequalities with respect to $\boldz$. 
 Furthermore, the truncation points in Theorem~\ref{theo1} can be explicitly stated as follows.
\begin{theo}
\label{theo3} 

Let $\theta \in [k\bar{k}]$ be the index of the first $k\bar{k}$ affine constraints in Eq.~\eqref{eq:selection_event} and $C := \{1,\ldots, k\bar{k}\}$. Furthermore, for notational simplicity, assume that first $k$ features are selected and remaining $\bar{k} = d - k$ features are unselected. Then, the marginal screening event in Eq.~\eqref{eq:selection_event} is written as
\begin{align*}
&(\boldA\boldz)_\theta = \widehat{{I}}(X_{\ell(\theta)},Y) - \widehat{{I}}(X_{m(\theta)},Y) \leq 0 \\
&~\text{with}~ m(\theta) := \left\lceil \theta/\bar{k} \right\rceil, \ell(\theta) := k + (\theta~ \textnormal{mod}~\bar{k})~\theta \in C.
\end{align*}

The lower and the upper truncation points for $m$-th feature are written as
\begin{align*}
&V^-(\boldA,\boldzero) \\
&\!:=\! \max_{\theta \in \calD }\!\left\{\! \frac{\! [\boldSigma]_{m,m}\! \!\left(\!\widehat{{I}}(X_{m(\theta)}, Y) \!-\! \widehat{{I}}(X_{\ell(\theta)}, Y)\!\right)\!}{[\boldSigma]_{\ell(\theta),m} - [\boldSigma]_{m(\theta),m}}\! \right\}  \!\!+\! \widehat{{I}}(X_{m}, Y), \\
&V^+(\boldA,\boldzero) \\ 
&\!:=\! \min_{\theta \in \bar{\calD}}\!\left\{\! \frac{\![\boldSigma]_{m,m} \!\!\left(\!\widehat{{I}}(X_{m(\theta)}, Y) \!-\! \widehat{{I}}(X_{\ell(\theta)}, Y)\!\right)\!}{[\boldSigma]_{\ell(\theta),m} - [\boldSigma]_{m(\theta),m}} \!\right\} \!\!+\! \widehat{{I}}(X_{m}, Y), 
\end{align*} 
where 
\begin{align*}
\calD &=  \{\theta ~~|~~ [\boldSigma]_{\ell(\theta),m} < [\boldSigma]_{m(\theta),m}\}, \\
\bar{\calD} &=  \{\theta ~~|~~ [\boldSigma]_{\ell(\theta),m} > [\boldSigma]_{m(\theta),m}\}.
\end{align*}
Proof: For deriving the lower and upper bounds, we simply need to plug  $\boldeta = \bolde_m$ (the unit vector whose $m$-th element is one and zero otherwise),  $\boldb = \boldzero$, $-(\boldA \boldz)_\theta  = \widehat{I}(X_{m(\theta)}, Y) - \widehat{I}(X_{\ell(\theta)}, Y)$, and $(\boldA \boldc)_\theta  = \left[\frac{[\boldSigma]_{\ell(\theta),m} - [\boldSigma]_{m(\theta),m}}{[\boldSigma]_{m,m}}\right]$ into Eq.~ \eqref{eq:lower_psi} and Eq.~\eqref{eq:upper_psi}. \proofend 
\end{theo}

\subsection{HSIC based Post Selection Inference (\texttt{hsicInf})}
Here, we propose the Hilbert-Schmidt Independence Criterion (HSIC) \citep{ALT:Gretton+etal:2005} based PSI algorithm.

\vspace{.1in}
\noindent {\bf Hilbert-Schmidt Independence Criterion (HSIC)}

HSIC is defined as \citep{ALT:Gretton+etal:2005}:
\begin{align}
\label{eq:HSIC}
\textnormal{HSIC}(X,Y) &\!=\! \mathbbE_{\boldx,\boldx',\boldy,\boldy'}[K(\boldx,\boldx')L(\boldy,\boldy')] \nonumber \\
&\phantom{=}+ \mathbbE_{\boldx,\boldx'}[K(\boldx,\boldx')]\mathbbE_{\boldy,\boldy'}[L(\boldy,\boldy')]  \\
&\phantom{=} - 2\mathbbE_{\boldx,\boldy}\left[\mathbbE_{\boldx'}[K(\boldx,\boldx')] \mathbbE_{\boldy'}[L(\boldy,\boldy')]\right], \nonumber
\end{align}
where $K(\boldx,\boldx')$ and $L(\boldy,\boldy')$ are unique positive definite kernel, and $\mathbbE_{\boldx,\boldx',\boldy,\boldy'}$ denotes the expectation over independent pairs $(\boldx,\boldy)$ and $(\boldx',\boldy')$ drawn from $p(\boldx,\boldy)$. With the use of characteristic kernels \citep{fukumizu2004dimensionality,sriperumbudur2011universality},  HSIC takes zero if $X$ and $Y$ are independent, and takes positive values otherwise. Thus,  we can select important features by ranking HSIC scores $\{\textnormal{HSIC}(X_m,Y)\}_{m = 1}^d$ in descending order, where $X_m$ is the random variable of the $m$-th feature.

\vspace{.1in}
\noindent {\bf Empirical Block Hilbert-Schmidt Independence Criterion (HSIC)}

Let us assume that the number of samples $n$ can be dis-jointly divided into $\frac{n}{B}$ blocks, where $B$ is the number of samples in each block. Then, the disjoint set of blocked paired samples is denoted by $\{\{(\boldx_i^{(b)}, \boldy_i^{(b)})\}_{i = 1}^{B}\}_{b = 1}^{n/B}$. 

An empirical estimate of the \emph{unbiased} block HSIC is given by \citep{zhang2016large}
\begin{align*}
&\widehat{\text{HSIC}}(X,Y) = \frac{B}{n} \sum_{b = 1}^{n/B} \widehat{\eta}_b,\\
&\widehat{\eta}_b = \frac{1}{B(B-3)} [\text{tr}(\bar{\boldK}^{(b)}\bar{\boldL}^{(b)}) + \frac{\boldone_B^\top \bar{\boldK}^{(b)}\boldone_B \boldone_B^\top \bar{\boldL}^{(b)}\boldone_B}{(B-1)(B-2)} \\ 
&\phantom{\widehat{\eta}_b =} - \frac{2}{B-2} \boldone_B^\top \bar{\boldK}^{(b)}\bar{\boldL}^{(b)} \boldone_B ],
\end{align*}
where $\boldK^{(b)} \in \mathbbR^{B\times B}$ is the input Gram matrix, $\boldL^{(b)} \in \mathbbR^{B\times B}$ is the output Gram matrix, $[\bar{\boldK}^{(b)}]_{ij} = [\boldK^{(b)}]_{ij} - \delta_{ij}[\boldK^{(b)}]_{ij}$ and $[\bar{\boldL}^{(b)}]_{ij} = [\boldL^{(b)}]_{ij} - \delta_{ij}[\boldL^{(b)}]_{ij}$, $\delta_{ij}$ takes 1 when $i = j$ and 0 otherwise, and $\boldone_B \in \mathbbR^{B}$ is the vector whose elements are all one. Since $\widehat{\eta}_b$ is computed from a partition of  i.i.d. samples from $p(\boldx,\boldy)$, $\{ \widehat{\eta}_b \}_{b=1}^{n/B}$ is also an i.i.d. random variables.

The empirical block HSIC score asymptotically follows normal distribution when $B$ is finite and $n$ goes to infinity, and thus, we can use the block HSIC for PSI based on Theorem~\ref{theo3}. Note that, to ensure Gaussian assumption, we need to have relatively large number of samples $n$ with a finite block size $B$.

\vspace{.1in}
\noindent {\bf  Choice of Kernel}:
For regression problems, we use the Gaussian kernel for both input and output as $\boldK^{(b)} \in \mathbbR^{B\times B}$ and $\boldL^{(b)} \in \mathbbR^{B\times B}$:
\begin{align*}
[\boldK^{(b)}]_{ij} &= \exp \left(-\frac{\|\boldx_i^{(b)} - \boldx_j^{(b)}\|^2_2}{2\tau_x^2}\right), \\
[\boldL^{(b)}]_{ij} &= \exp \left(-\frac{\|\boldy_i^{(b)} - \boldy_j^{(b)}\|^2_2}{2\tau_y^2}\right), 
\end{align*}
where $\tau_x > 0$ and $\tau_y > 0$ are kernel parameters. 

For $L$-class classification problems (i.e., $y \in \{1,2,\ldots,L\}$), we can first transform the target vector $y = \ell$ as
\begin{align*}
\boldy_j &= [{0~\cdots 0~\underbrace{1}_{\ell}~0~\cdots 0}]^\top \in \mathbbR^{L}, 
\end{align*}
and use the linear kernel:
\begin{eqnarray*}
[\boldL^{(b)}]_{ij} = \boldy_{i}^\top \boldy_{j}  = \left\{
\begin{array}{cc}
1& (y_i = y_j)\\
0& \textnormal{Otherwise}
\end{array}
\right..
\end{eqnarray*}
Note that, this is equivalent to the delta kernel \citep{song2012feature}.

\vspace{.1in}
\noindent {\bf Mean and covariance matrix estimation:}
Suppose that the mean and variance of the within-block estimator $\widehat{\eta}_b$ are $\widetilde{\mu}$ and $\widetilde{\sigma}$, respectively.  Then, the vector of empirical HSICs $[\widehat{\textnormal{HSIC}}(X_1,Y), \ldots, \widehat{\textnormal{HSIC}}(X_d,Y)]^\top \in \mathbbR^{d}$ converges in distribution to a multi-variate normal by the central limit theorem, where the mean and covariance matrices are given by by $\widetilde{\mu}$ and $n/B\widetilde{\sigma}$, respectively. 
In estimating $[\widetilde{\boldSigma}]_{m,m'}$, the standard covariance estimator is applied to the ($n/B$) within-block estimators for $m$ and $m'$.  Note that, when $n/B$ is too small, we can use a high-dimensional covariance estimation algorithm such as POET \citep{fan2013large}.

\vspace{.1in}
\noindent {\bf Post Selection Inference:}
We consider the following hypothesis tests:
\begin{itemize}
\item $H_{0,m}$: $\textnormal{HSIC}(X_m, Y) = 0 ~|~   \text{$\calS$ was selected}$, 
\item $H_{1,m}$: $\textnormal{HSIC}(X_m, Y)~ \neq 0 ~|~  \text{$\calS$ was selected}$.
\end{itemize} 
 
Then, the $p$-value of the $m$-th feature is estimated by using  Theorem~\ref{theo3}. 


\section{Related Work}
In this section, we briefly review related works. It has long been recognized that \emph{selection bias} must be corrected for statistical inference after feature selection. In machine learning community, the most common approach for dealing with selection bias is data splitting. In data splitting, the dataset is divided into two disjoint sets, and one of them is used for feature selection, and the other is used for statistical inference. Since the inference phase is made independently of the feature selection phase, we do not have to care about the selection effect. An obvious drawback of data splitting is that the powers are low both in feature selection and inference phases. Since only a part of the dataset can be used in feature selection phase, the risk of failing to select truly important features would increase. Similarly, the power of statistical inference (i.e., the probability of true positive finding) would decrease because the inference is made with a smaller dataset. In addition, it is quite annoying that different features might be selected if the dataset is divided differently. It is important to note that data splitting is also regarded as a selective inference because the inference is made only for the selected features in the feature selection phase, and the other unselected features are ignored.

In statistics, \emph{simultaneous inference} has been studied traditionally for selection bias correction. For feature selection bias correction, all possible subsets of features are considered. Let $\widehat{\cal S}$ represent the set of selected features and $T_j({\cal S})$ be a test statistic for the $j^{\rm th}$ selected feature in $\widehat{\cal S}$. In simultaneous inference, critical points $l$ and $u$ at level $\alpha$ are determined to satisfy
\begin{align}
 \label{eq:simultaneous-inference1}
 P(T_j(\widehat{\cal S}) \notin [l, u] \text{ for any subset of features $\widehat{\cal S}$}) \le \alpha. 
\end{align}
The probability in \eqref{eq:simultaneous-inference1} is also written as 
\begin{align*}
 &P(T_j(\widehat{\cal S}) \notin [l, u] \text{ for any subset of features $\widehat{\cal S}$}) \\ 
 &= \sum_{{\cal S}} P(T_j(\widehat{\cal S}) \notin [l, u] \mid \widehat{\cal S} = {\cal S}) P(\widehat{\cal S} = {\cal S}),
\end{align*}
where the summation in the right hand side runs over all possible subsets of features. Unfortunately, unless the number of original features is fairly small, it is computationally challenging to consider all possible subsets of features $\widehat{{\cal S}}$ \citep{berk2013valid}. 

In selective inference, we only consider the case that a certain $\cal S$ is selected, and we determine the critical points $l$ and $u$ so that \emph{selective type I error} is controlled, i.e., 
\begin{align}
 \label{eq:selective-inference}
 P(T_j(\widehat{\cal S}) \notin [l, u] \mid \widehat{\cal S} = {\cal S}) \le \alpha. 
\end{align}
Selective inference framework in the form of \eqref{eq:selective-inference} has been increasingly popular after the seminal work by \cite{lee2016exact}. In this work, the authors studied selective inference after a subset of features are selected by LASSO \citep{JRSSB:Tibshirani:1996}. Their novel finding is that, in linear regression models with Gaussian noise, if the selection event can be represented by a set of linear inequalities with respect to the response variables (as in LASSO case), then any linear combination of the responses conditioned on the selection event is distributed according to some truncated normal distribution as discussed in Theorem 1. This seminal result is known as a polyhedral lemma and is very useful for deriving a null distribution of a test statistic in the context of selective inference. After this work, selective inference framework has been studied for several problems where the assumptions of polyhedral lemma are satisfied (see, e.g., \citep{lee2014exact}). 

Unfortunately, polyhedral lemma in \citep{lee2016exact} can be used only when the responses are normally distributed. It is thus difficult to generalize selective inference framework to other important problem such as classification, multi-task learning and so on. To the best of our knowledge, there are only a few attempts of selective inference in those generalized settings \citep{taylor2016post}. The idea in these attempts is to use asymptotic theory, but the underlying assumption used in these studies is somewhat restrictive. Specifically, they require that at least one truncation points are bounded in probability tending to 1, which is somewhat embarrassing because the truncation points are not bounded in the case of classical inference without feature selection. In contrast, our proposed method is not suffered from such a restriction because we only use an asymptotic normality of $\widehat{\rm HSIC}(X_j, Y)$. In our method, since the test is conducted within the reproducing kernel Hilbert space, selective inference is possible for a variety of response types that includes classification, ranking, or more generally, any structures responses.

\section{Experiment} 
In this section, we evaluate the proposed algorithm in regression and classification problems.
 
\subsection{Setup}
We compared the performance of the proposed methods with the linear PSI method based on the least angle regression \citep{lee2016exact,efron2004least}, which is a state-of-the-art PSI algorithm.  In this paper, we used the \texttt{larInf} function in the R package \texttt{selectiveInference}. Moreover, we compared the proposed method with \texttt{hsic}, which first selects features by empirical HSIC and then test each HSIC score without adjusting the sampling distribution (i.e., $V^{-}(\boldA,\boldzero) = -\infty$ and $V^{+}(\boldA,\boldzero) = \infty$), and the data-splitting approach (\texttt{split}). For both proposed and the existing methods, we set the number of selected features as $k = 10$ and the significance level $\alpha = 0.05$. For HSIC based approaches, we used the block parameter $B = \{5,10\}$. 

In PSI frameworks, the covariance matrix $\boldSigma$ is assumed to be known. However,  since the true covariane matrix is not available in practice,  we need to estimate the covariance matrix from data. To this end, for \texttt{hsicInf} and \texttt{hsic}, we divided the samples into \emph{two} disjoint sets; we used $\frac{n}{3}$ samples for estimating $\boldSigma$ and the rest of $\frac{2n}{3}$ samples for selecting features and computing HSIC. For \texttt{split}, we divided the samples into \emph{three} disjoint sets with sample size $\frac{n}{3}$. Then, we used  each of them for estimating $\boldSigma$, selecting and testing features, and computing HSIC, respectively. In this paper, we employed the POET algorithm for the covariance matrix estimation \citep{fan2013large}.


In regression setup, the kernel parameters are experimentally set to $(\tau_x,\tau_y) = (1.0,1.0)$ for uni-variate setup and $(\tau_x, \tau_y) = (1.0, \textnormal{med}(\{\|\boldy_i - \boldy_j\|_2\}_{i,j=1}^n))$ for multi-variate output, respectively, where each
feature was normalized to have mean zero and standard deviation 1. In classification setup, we used the Gaussian kernel for input and the delta kernel for output.  We reported the true positive rate (TPR) $\frac{k^\prime}{k^*}$ where $k^\prime$ is the number of truely relevant features that are reported to be positive, while $k^*$ is the number of truly relevant features. We further computed the false positive rate (FPR) $\frac{k^{\prime\prime}}{k}$ where $k^{\prime\prime}$ is the number of truely irelevant features that are falsely reported to be positive.  

\subsection{False positive rate control} 
First, to check whether the methods can properly control the desired FPR, we run the proposed methods using a dataset that has no relationship between input and output. Specifically, we generated the input output pairs as  $\{(\boldx_i, y_i)\}_{i = 1}^n$, where  $\boldx \sim N(\boldzero, \boldI_{20})$, $\boldzero \in \mathbbR$ is the vector whose elements are all zero, $\boldI \in \mathbbR^{20 \times 20}$ is the identity matrix, and $y \sim N(0,1)$.

\begin{figure*}[t!]
\begin{center}
\begin{minipage}[t]{0.45\linewidth}
\centering
  {\includegraphics[width=0.99\textwidth]{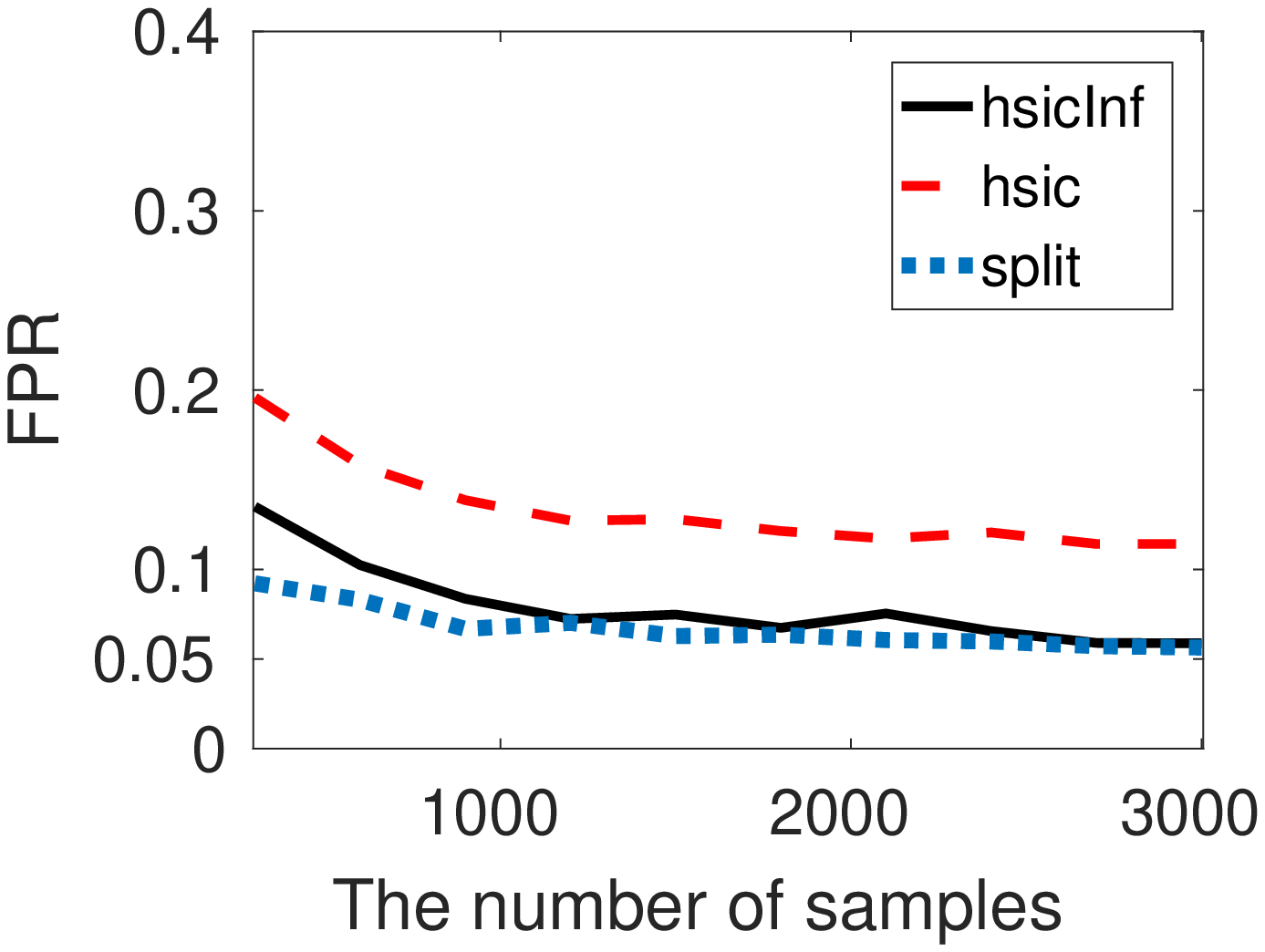}} \\ \vspace{-0.10cm}
(a) 
\end{minipage}
\begin{minipage}[t]{0.45\linewidth}
\centering
  {\includegraphics[width=0.99\textwidth]{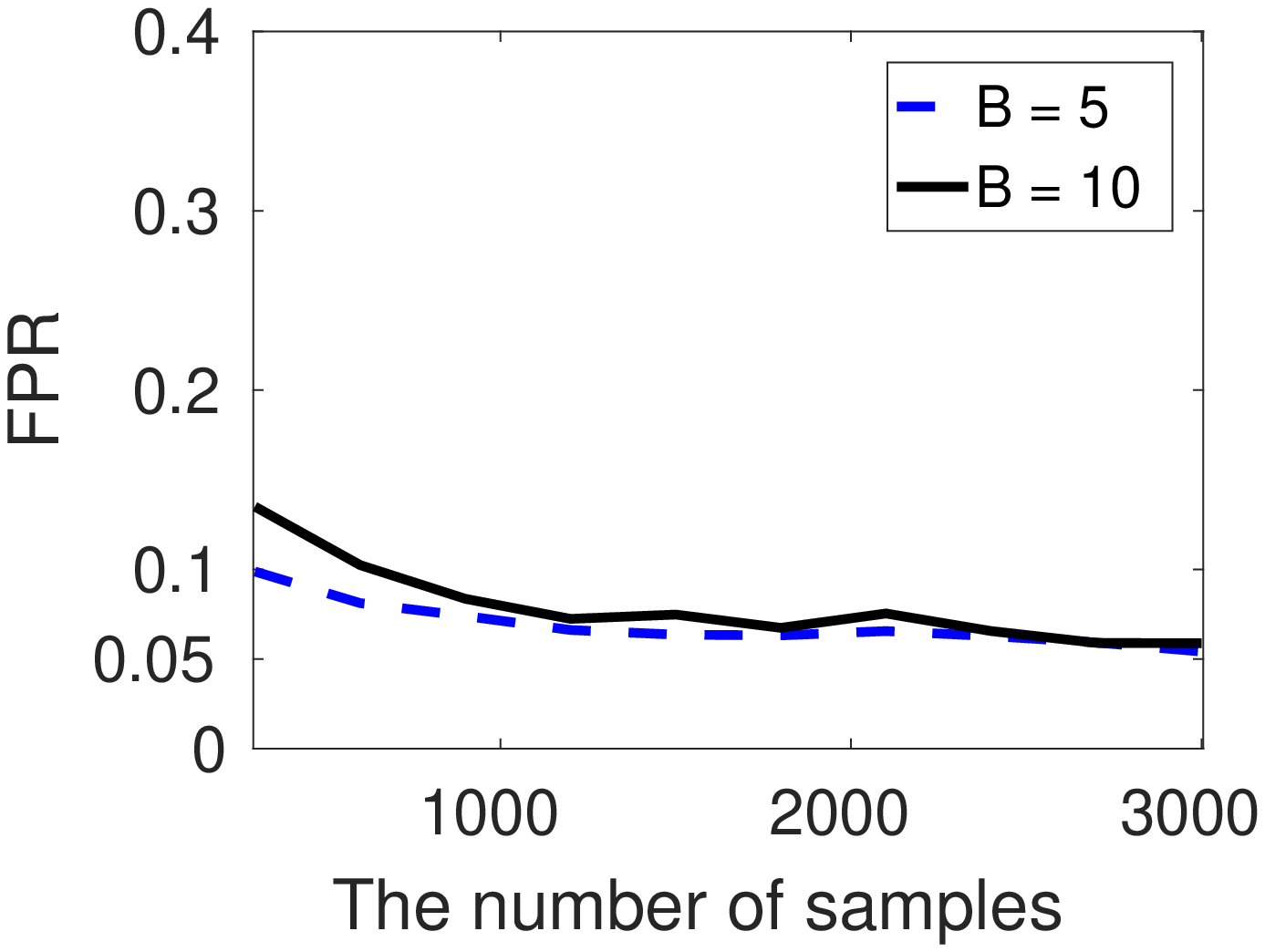}} \\ \vspace{-0.10cm}
(b) 
\end{minipage}
 \caption{False positive rates at significant level $\alpha = 0.05$ of the proposed methods. (a): Comparison of \texttt{hsicInf}, \texttt{hsic}, \texttt{split}, and \texttt{larInf}. We used $B=10$. (b): FPRs for \texttt{hsicInf} with different block parameter $B$. The \texttt{hsic}  computes $p$-values without adjusting the sampling distribution by  Theorem~\ref{theo3}.  }
    \label{fig:fpr}
\end{center}
\end{figure*}

Figure~\ref{fig:fpr}(a) shows that the FPRs of \texttt{hsicInf}, \texttt{hsic}, and \texttt{split} algorithms, respectively. The both the proposed method and \texttt{split} successfully control FPR, while \texttt{hsic} fails to control FPR. Thus, the adjustment of the sampling distribution is critical for estimating proper $p$-values.  It shows that all FPRs tend to be high when the number of samples are small, and gradually converging to the significance level when the number of samples increases. Figures~\ref{fig:fpr}(b) shows the FPRs for the \texttt{hsicInf} algorithms with different block size $B$. 

Since \texttt{hsic} cannot control the FPR at the desired level, we do not compare the TPR of \texttt{hsic} in the following section.

 \begin{figure*}[t!]
\begin{center}
\begin{minipage}[t]{0.325\linewidth}
\centering
  {\includegraphics[width=0.99\textwidth]{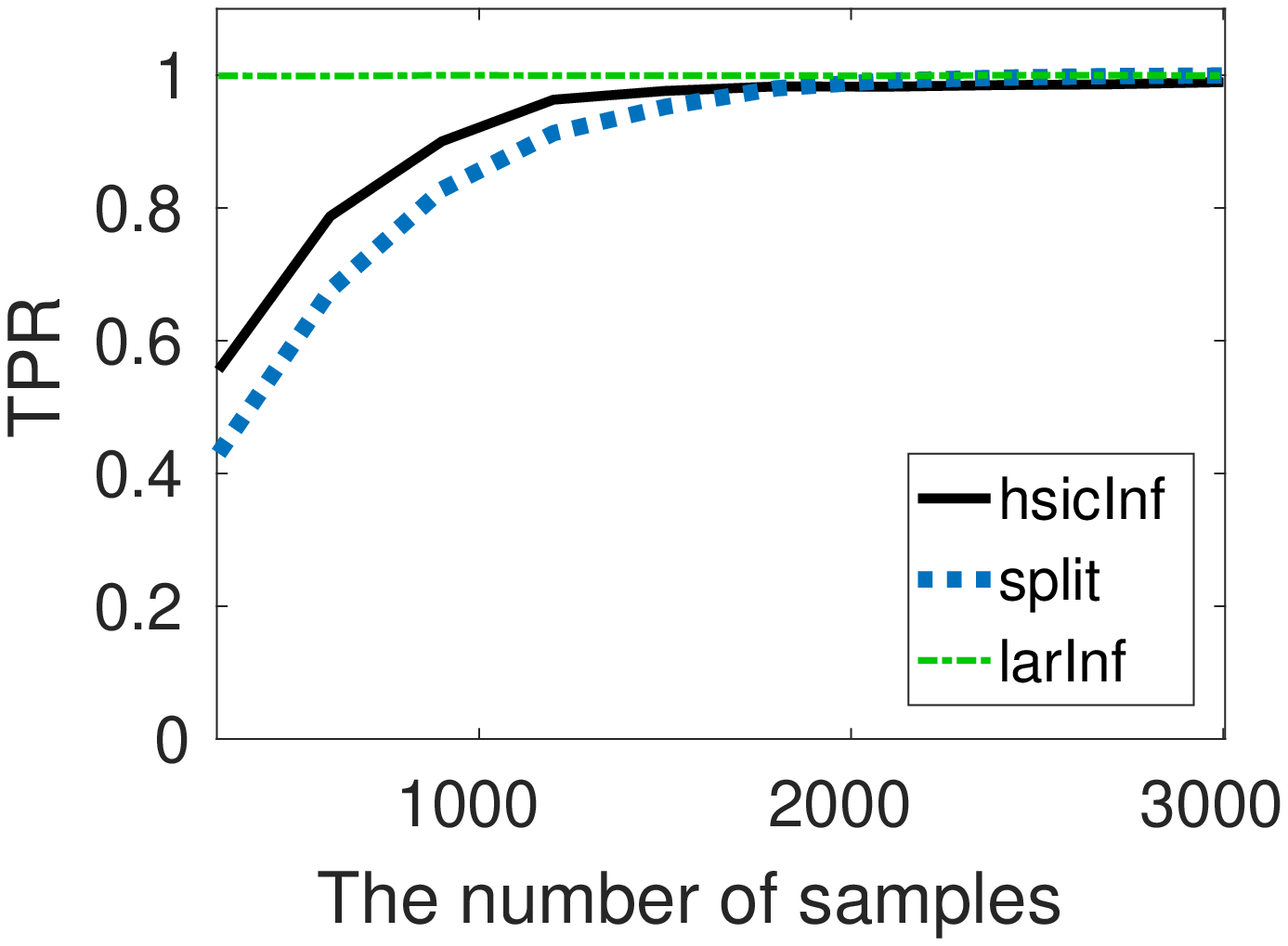}} \\ \vspace{-0.10cm}
(a) Linear.\vspace{0.2cm}
\end{minipage}
\begin{minipage}[t]{0.325\linewidth}
\centering
  {\includegraphics[width=0.99\textwidth]{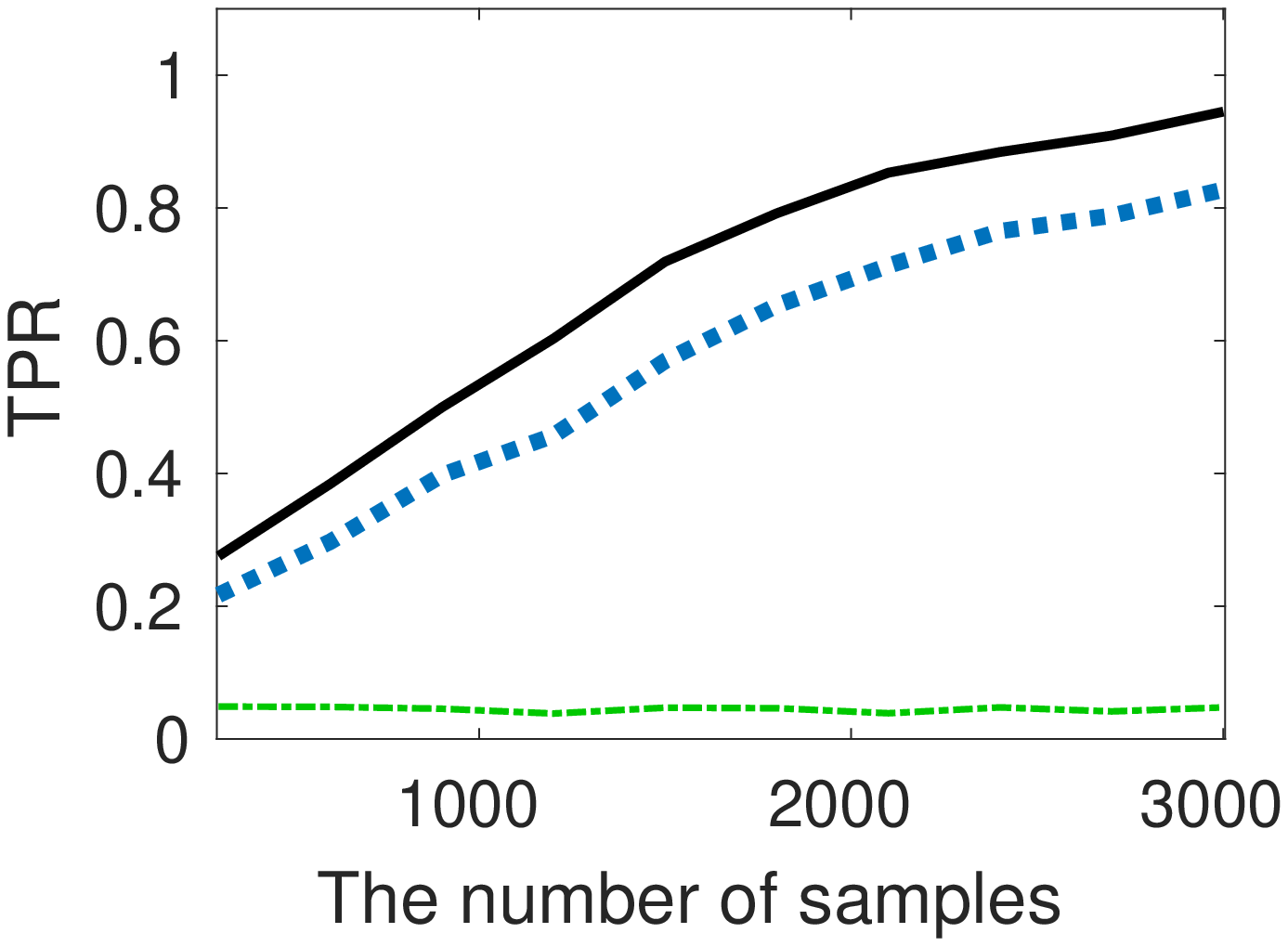}} \\ \vspace{-0.10cm}
(b) Additive Non-linear.\vspace{0.2cm}
\end{minipage}
\begin{minipage}[t]{0.325\linewidth}
\centering
  {\includegraphics[width=0.99\textwidth]{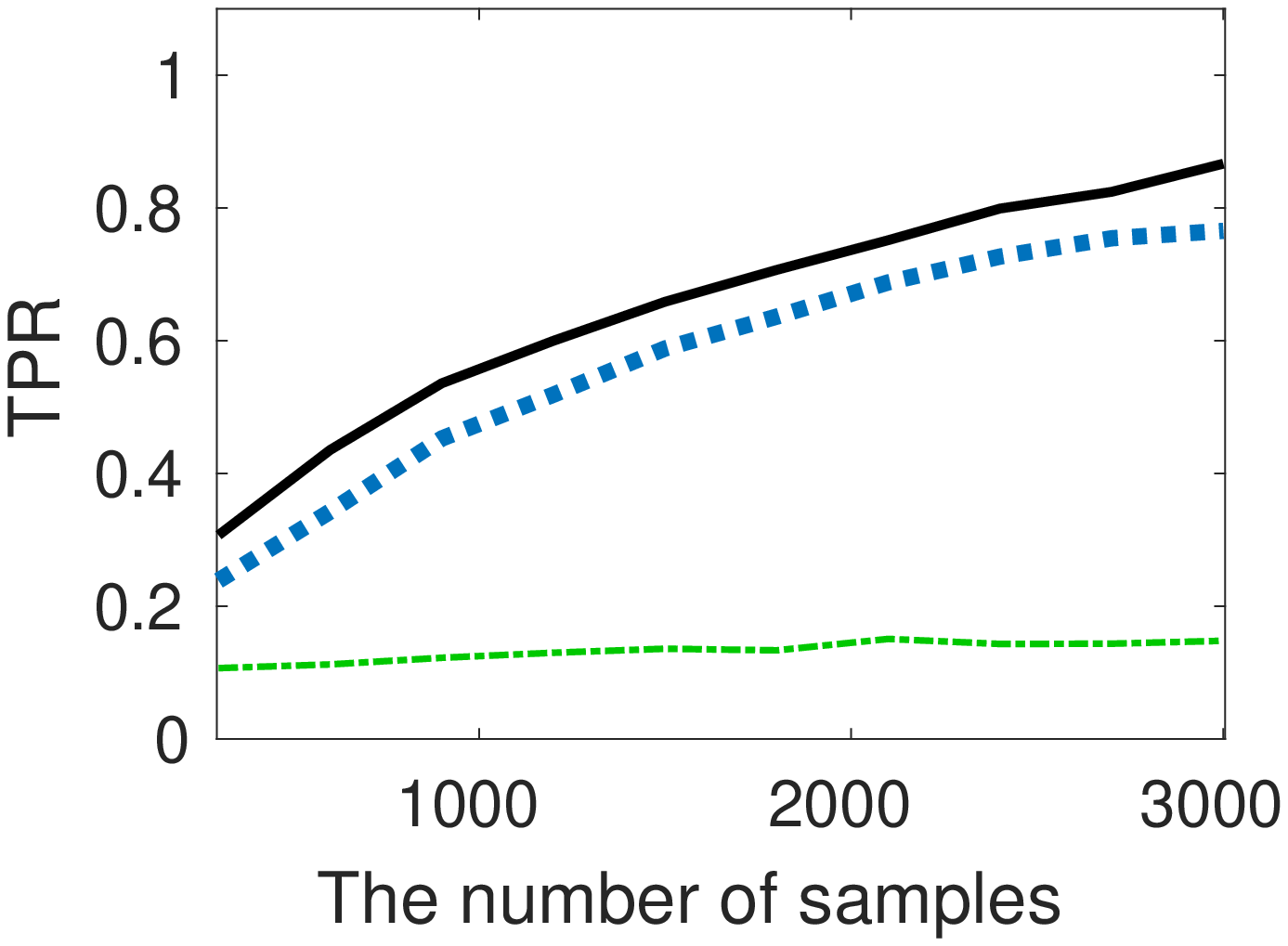}} \\ \vspace{-0.10cm}
(c) Non-additive Non-linear.\vspace{0.2cm}
\end{minipage}\\
\begin{minipage}[t]{0.325\linewidth}
\centering
  {\includegraphics[width=0.99\textwidth]{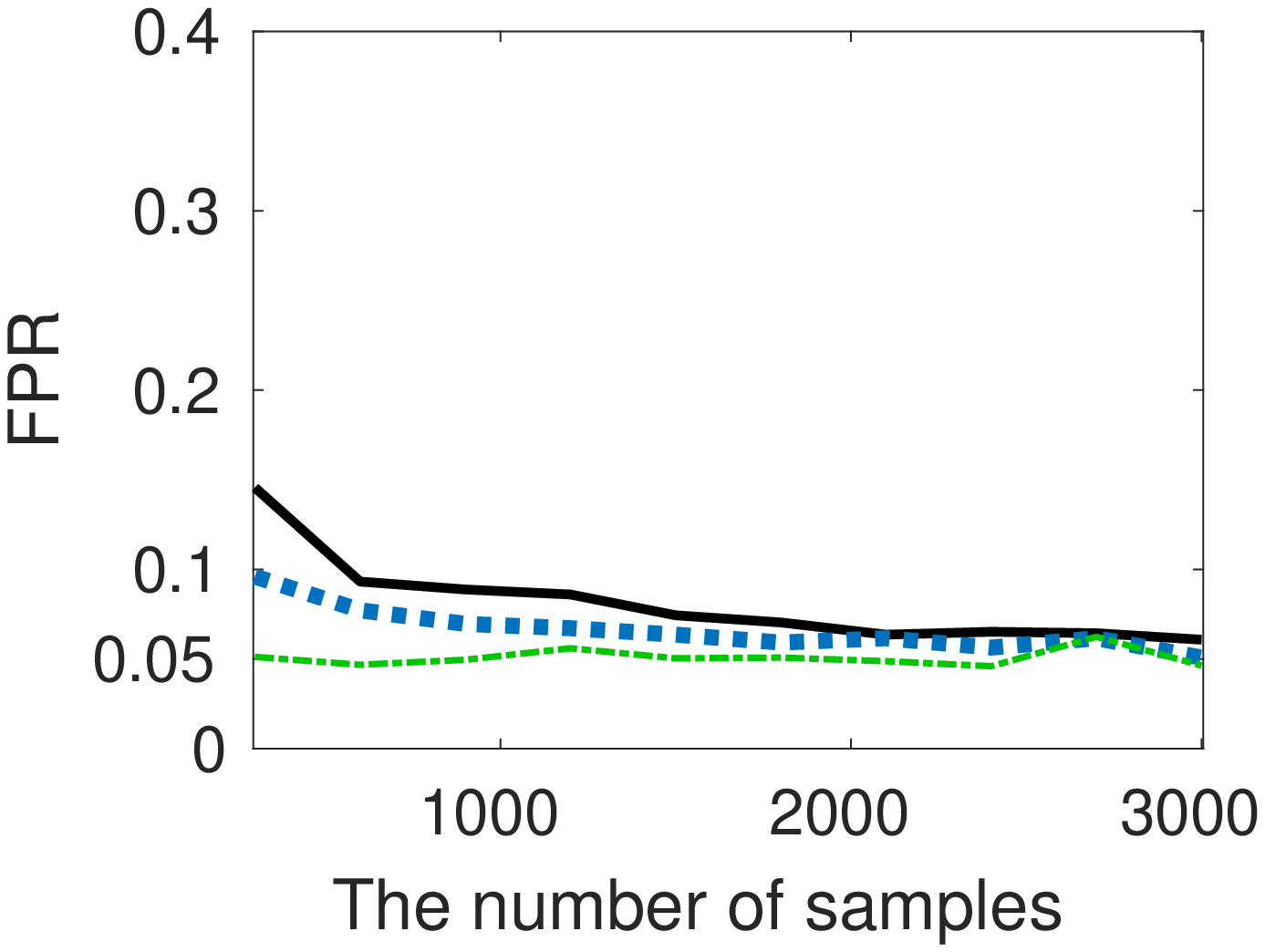}} \\ \vspace{-0.10cm}
(d) Linear.
\end{minipage}
\begin{minipage}[t]{0.325\linewidth}
\centering
  {\includegraphics[width=0.99\textwidth]{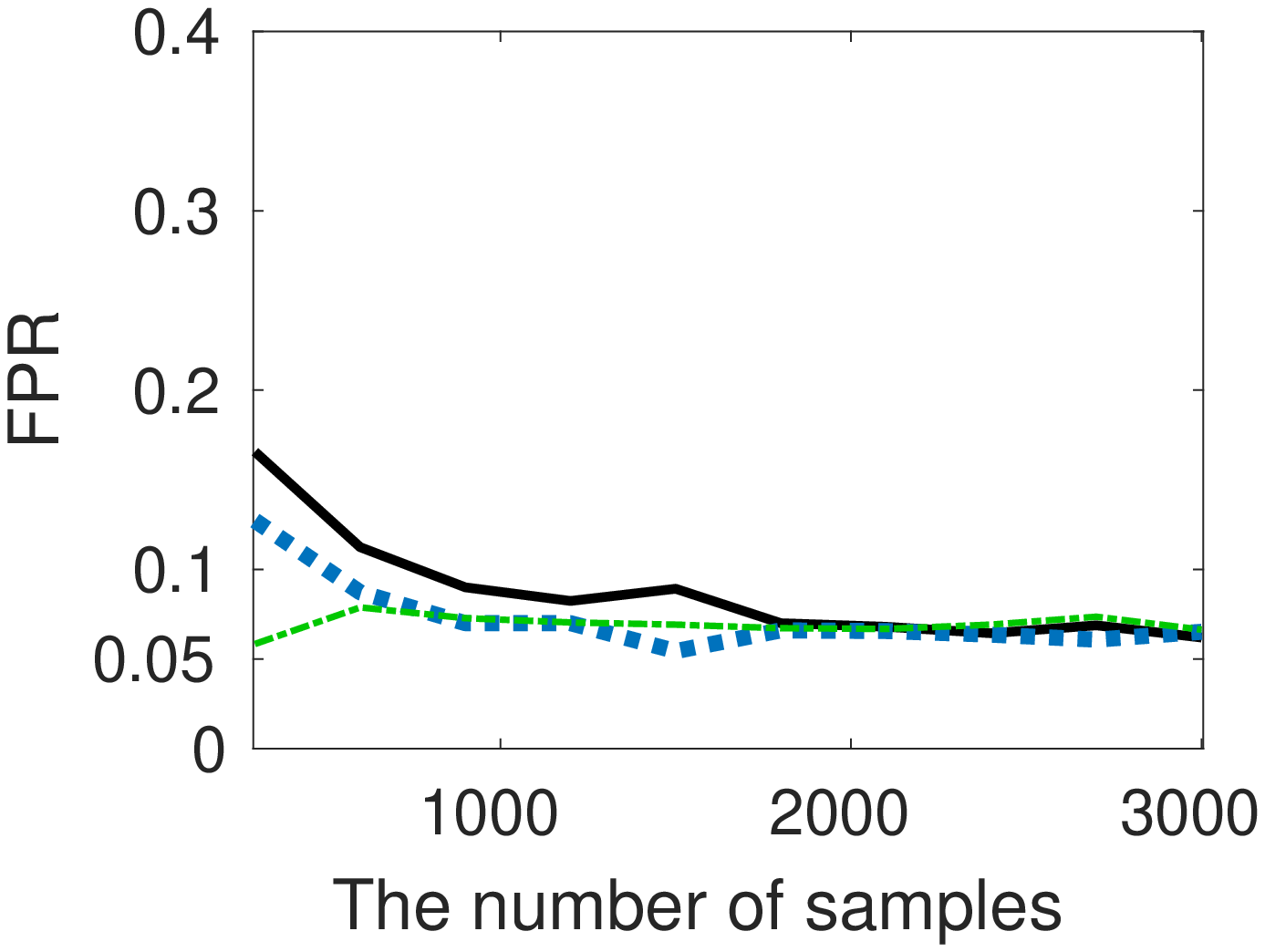}} \\ \vspace{-0.10cm}
(e) Additive Non-linear.
\end{minipage}
\begin{minipage}[t]{0.325\linewidth}
\centering
  {\includegraphics[width=0.99\textwidth]{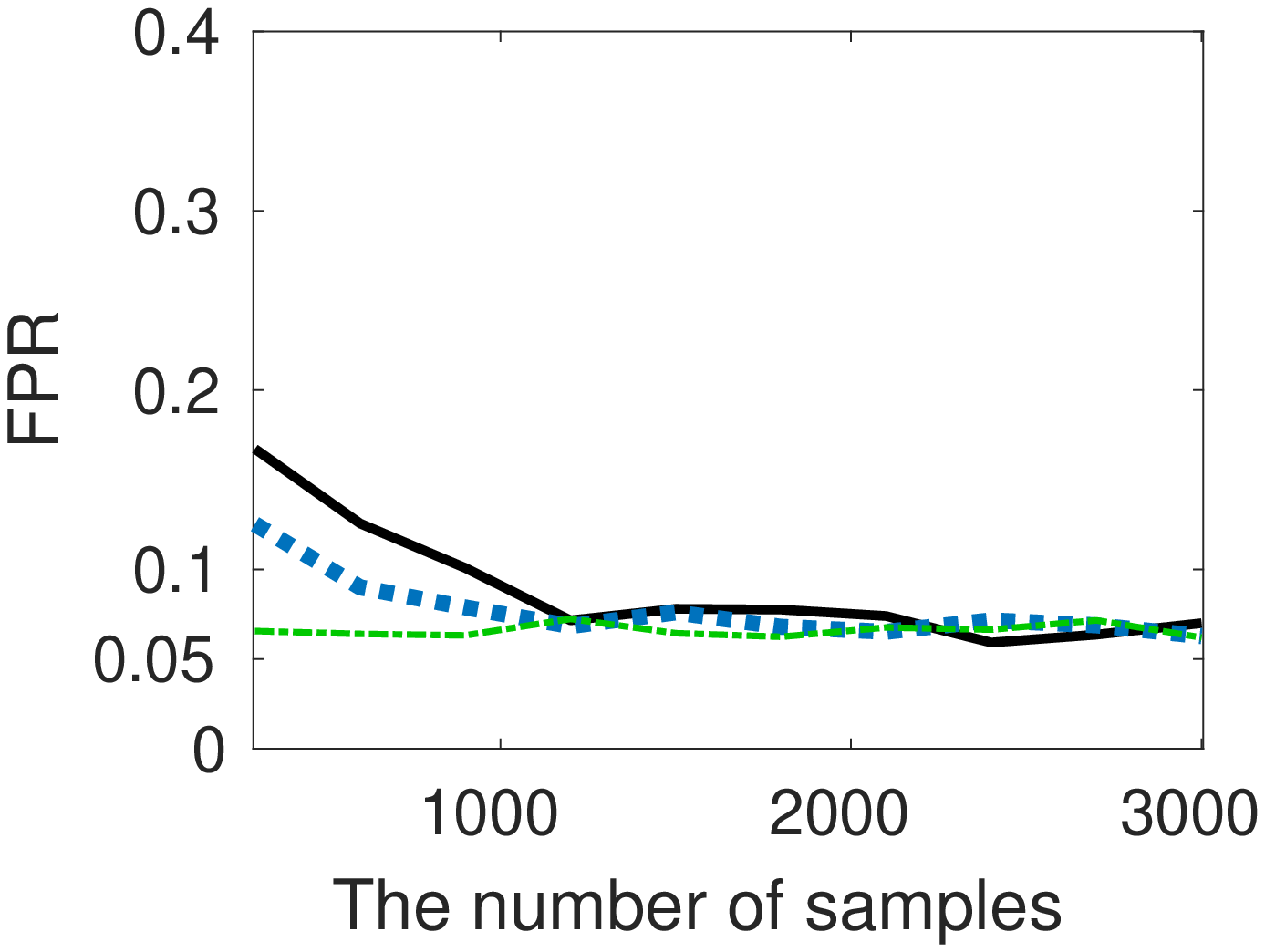}} \\ \vspace{-0.10cm}
(f) Non-additive Non-linear.
\end{minipage}\\
 \caption{The results for the uni-variate setups. We used $B = 10$ for the hsic related methods. (a)-(c): TPRs. (d)-(f): FPRs. }
    \label{fig:synth1}
\end{center}
\end{figure*}

 \begin{figure*}[t!]
\begin{center}
\begin{minipage}[t]{0.325\linewidth}
\centering
  {\includegraphics[width=0.99\textwidth]{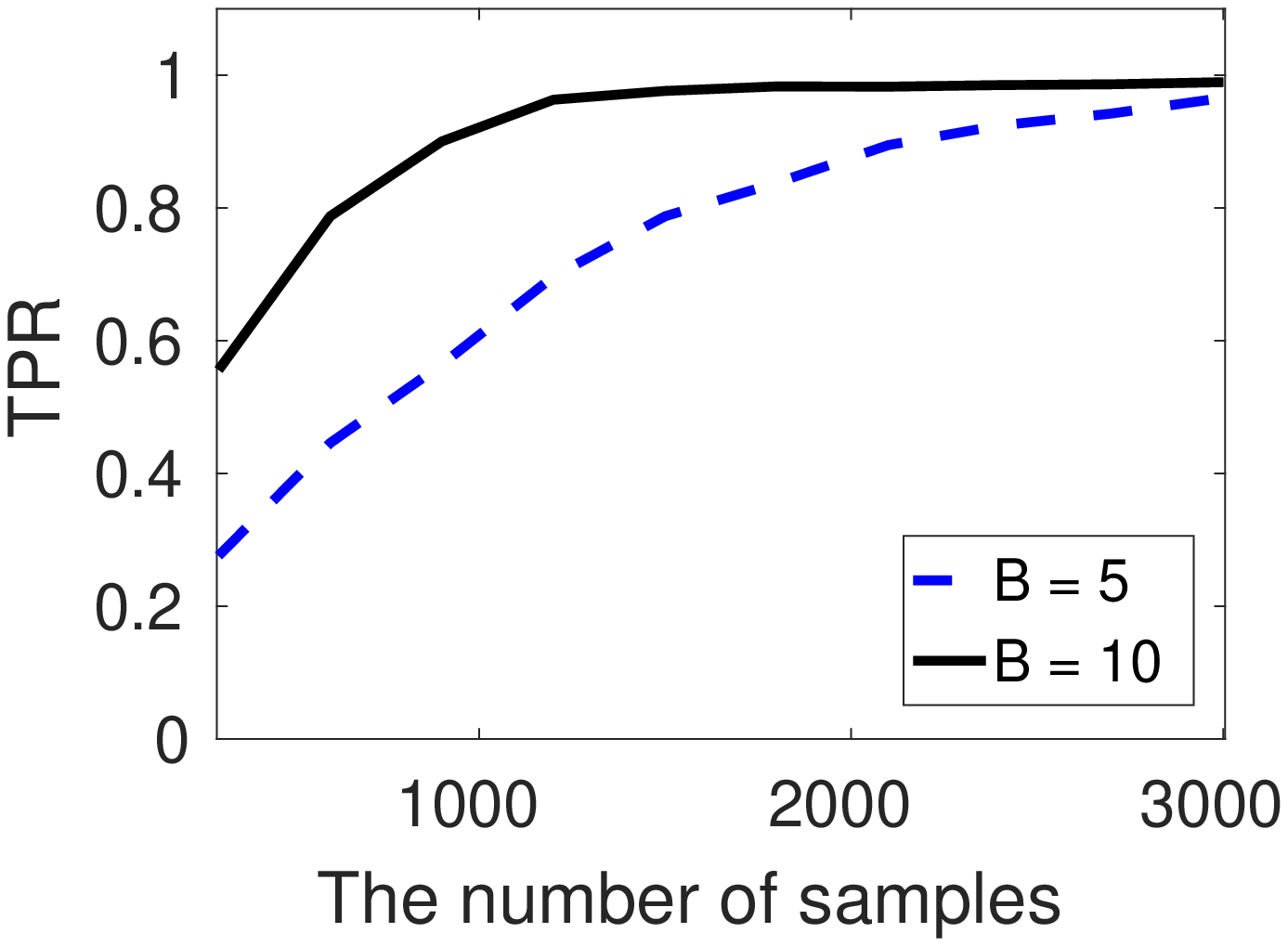}} \\ \vspace{-0.10cm}
(a) Linear.\vspace{0.2cm}
\end{minipage}
\begin{minipage}[t]{0.325\linewidth}
\centering
  {\includegraphics[width=0.99\textwidth]{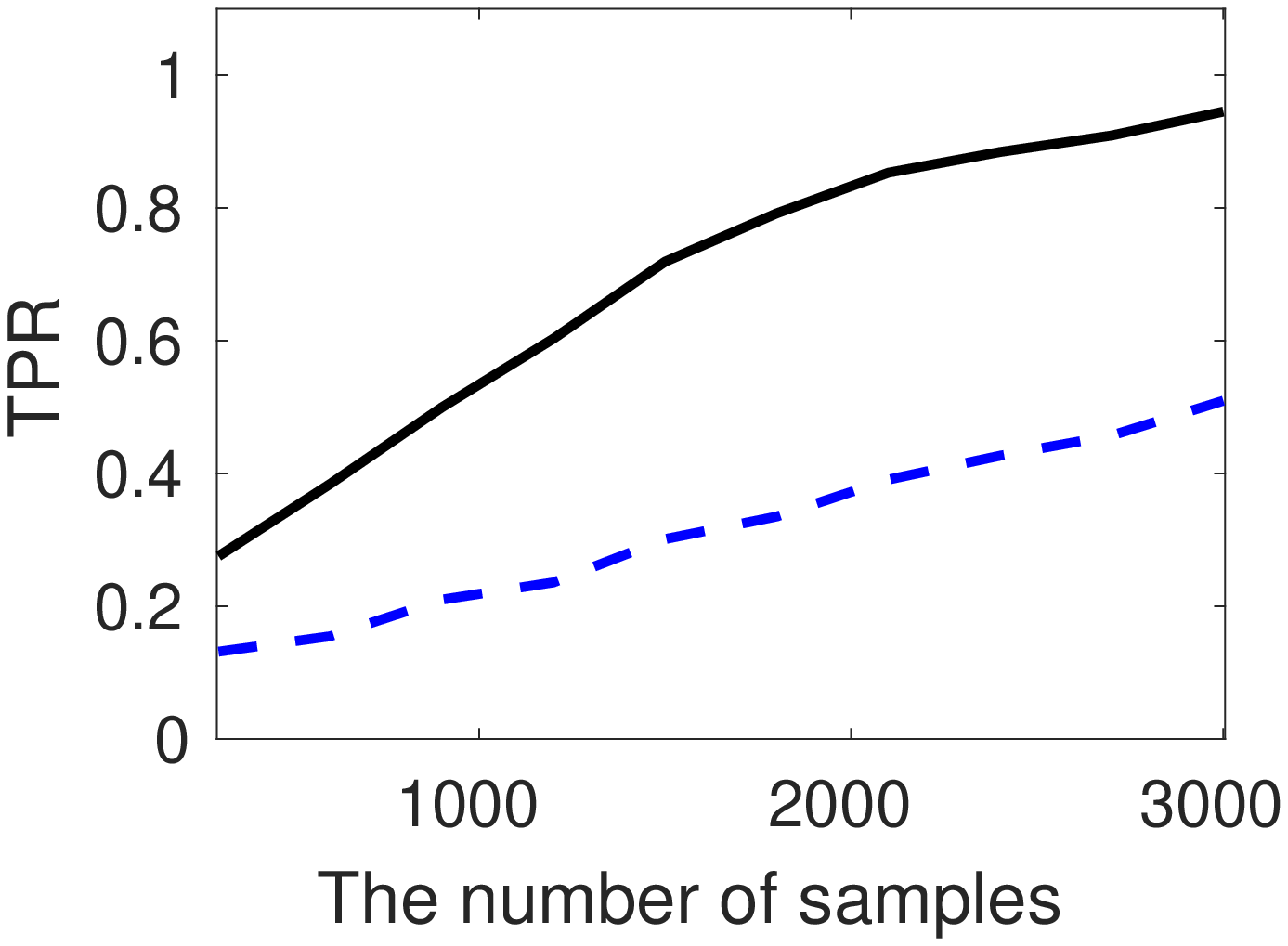}} \\ \vspace{-0.10cm}
(b) Additive Non-linear.\vspace{0.2cm}
\end{minipage}
\begin{minipage}[t]{0.325\linewidth}
\centering
  {\includegraphics[width=0.99\textwidth]{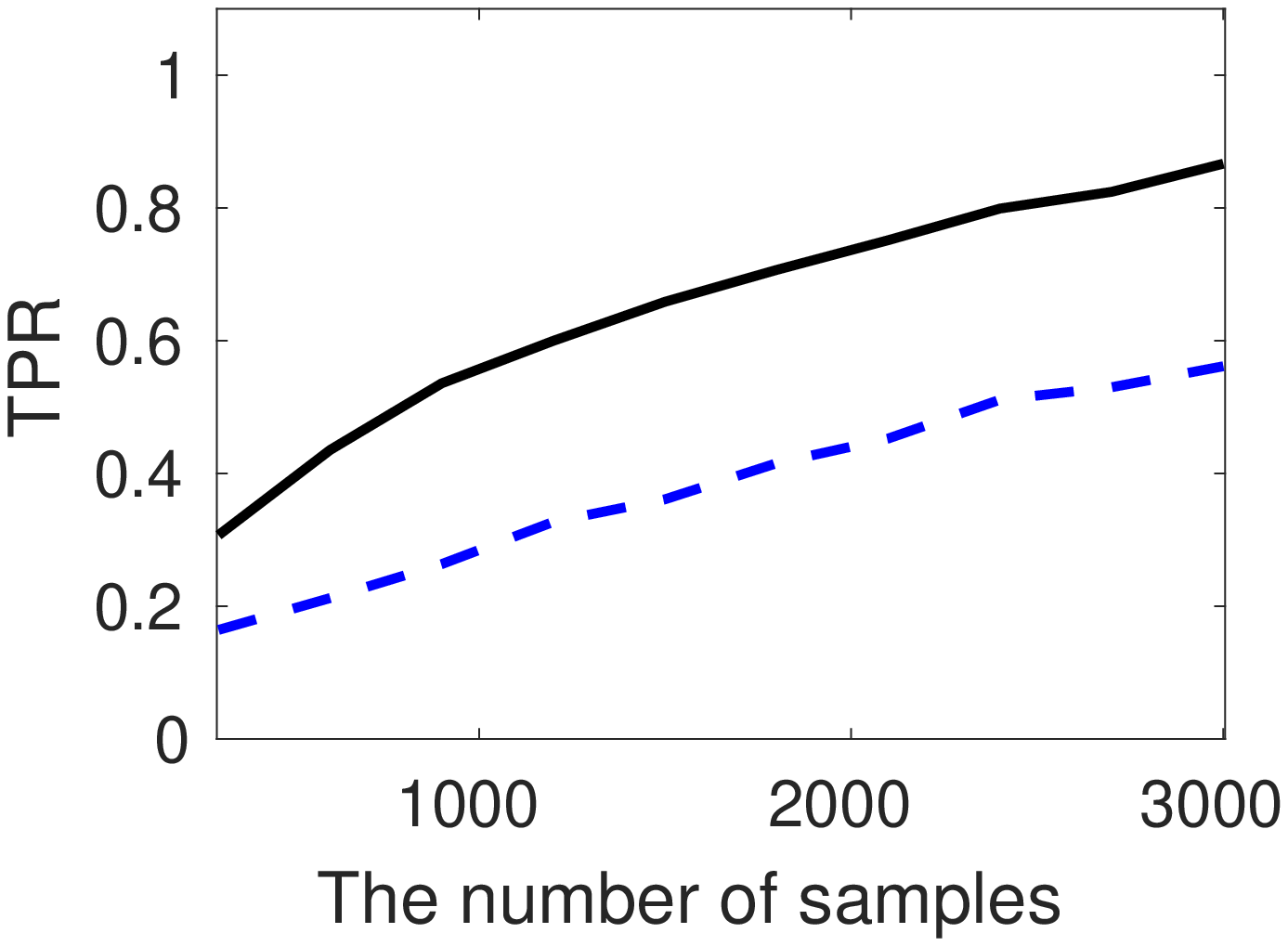}} \\ \vspace{-0.10cm}
(c) Non-additive Non-linear.\vspace{0.2cm}
\end{minipage}\\
\begin{minipage}[t]{0.325\linewidth}
\centering
  {\includegraphics[width=0.99\textwidth]{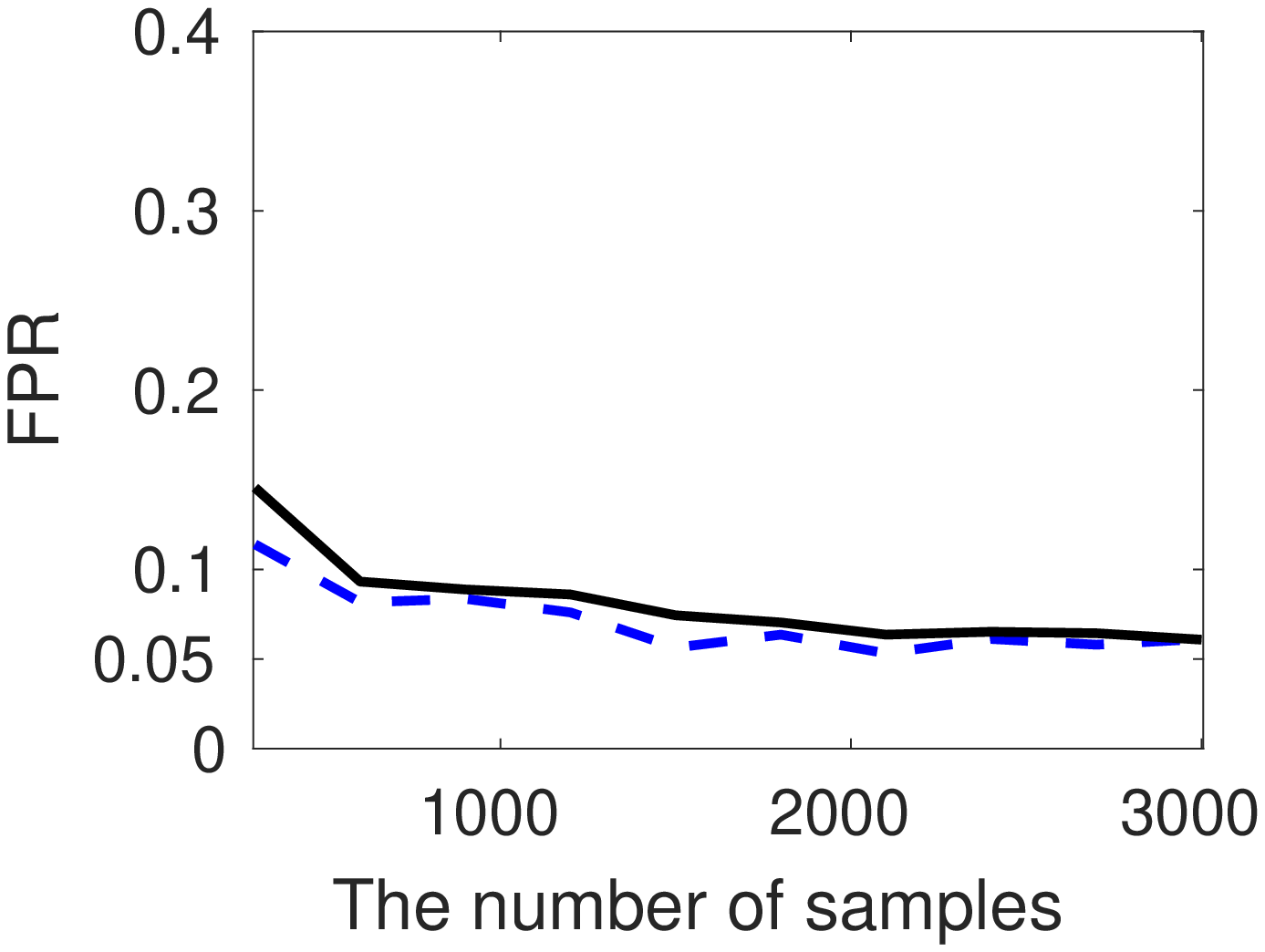}} \\ \vspace{-0.10cm}
(d) Linear.
\end{minipage}
\begin{minipage}[t]{0.325\linewidth}
\centering
  {\includegraphics[width=0.99\textwidth]{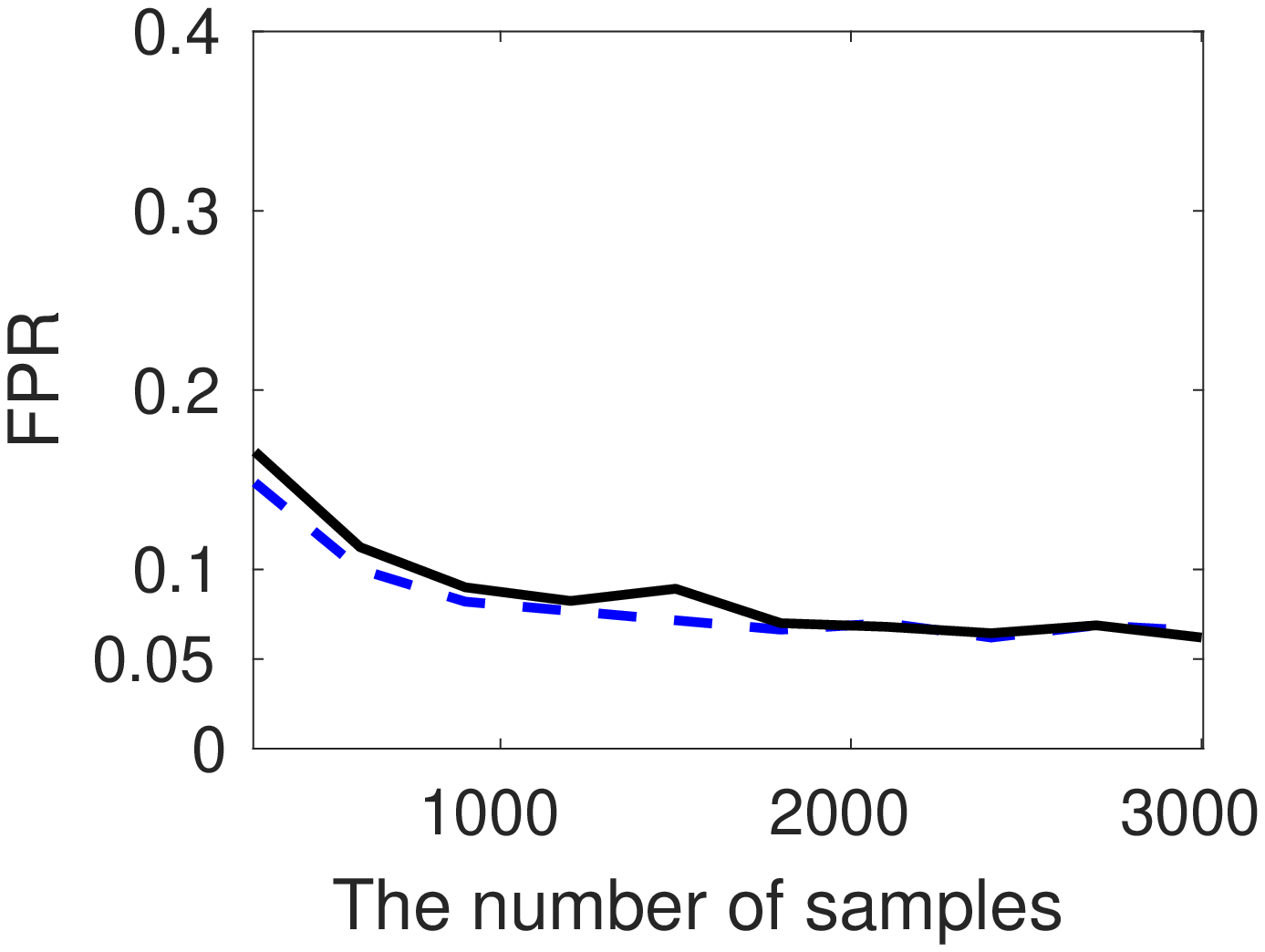}} \\ \vspace{-0.10cm}
(e) Additive Non-linear.
\end{minipage}
\begin{minipage}[t]{0.325\linewidth}
\centering
  {\includegraphics[width=0.99\textwidth]{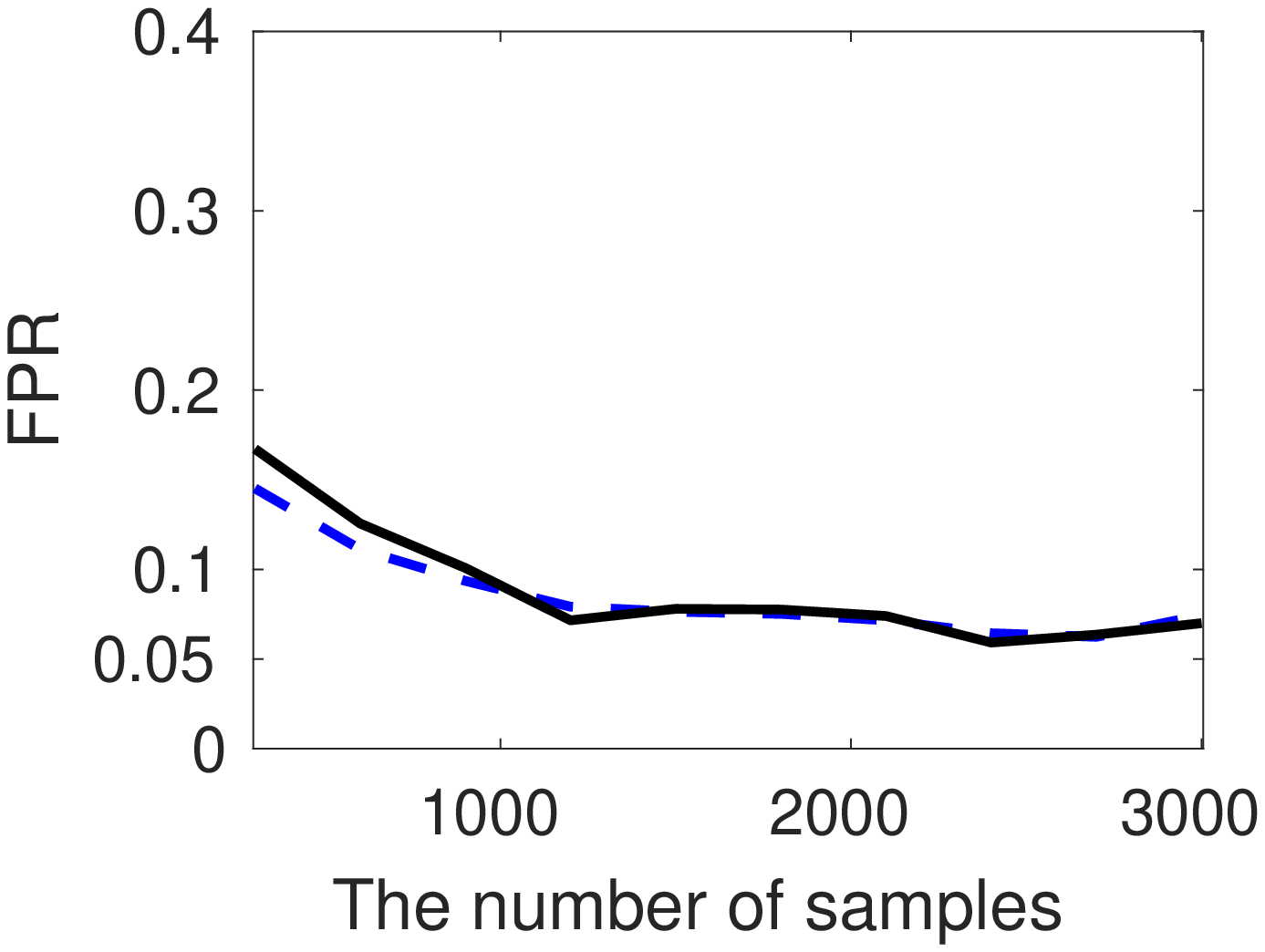}} \\ \vspace{-0.10cm}
(f) Non-additive Non-linear.
\end{minipage}\\
 \caption{The results for \texttt{hsicInf} in uni-variate setups with different block parameter $B$. (a)-(c): TPR for the three datasets. (d)-(f): FPR for the three datasets. }
    \label{fig:synth1_b_rank}
\end{center}
\vspace{-.2in}
\end{figure*}

\begin{figure*}[t!]
\begin{center}
\begin{minipage}[t]{0.245\linewidth}
\centering
  {\includegraphics[width=0.99\textwidth]{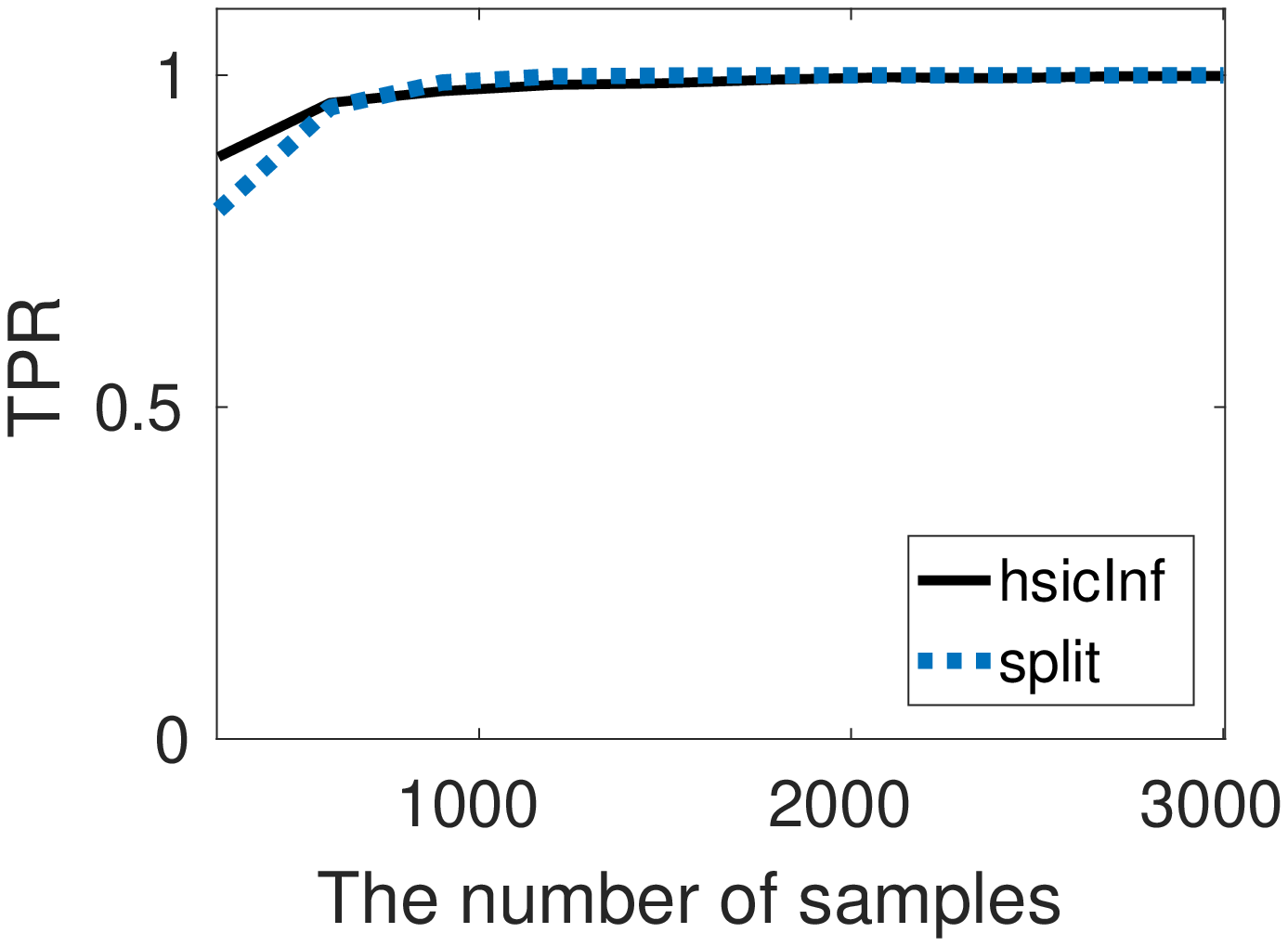}} \\ \vspace{-0.10cm}
(a) Multi-variate (TPR).
\end{minipage}
\begin{minipage}[t]{0.245\linewidth}
\centering
  {\includegraphics[width=0.99\textwidth]{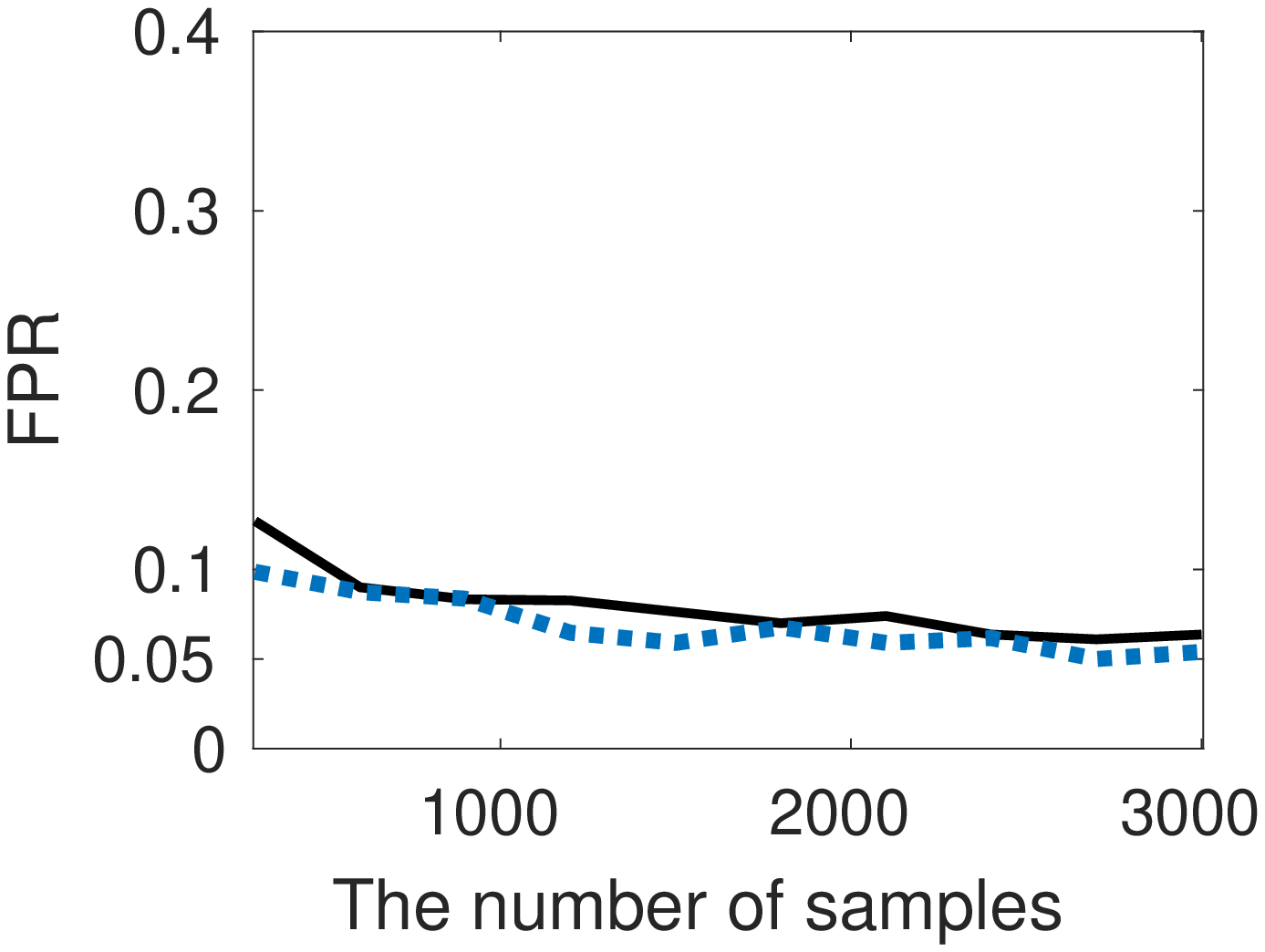}} \\ \vspace{-0.10cm}
(b) Multi-variate (FPR).
\end{minipage} 
\begin{minipage}[t]{0.245\linewidth}
\centering
  {\includegraphics[width=0.99\textwidth]{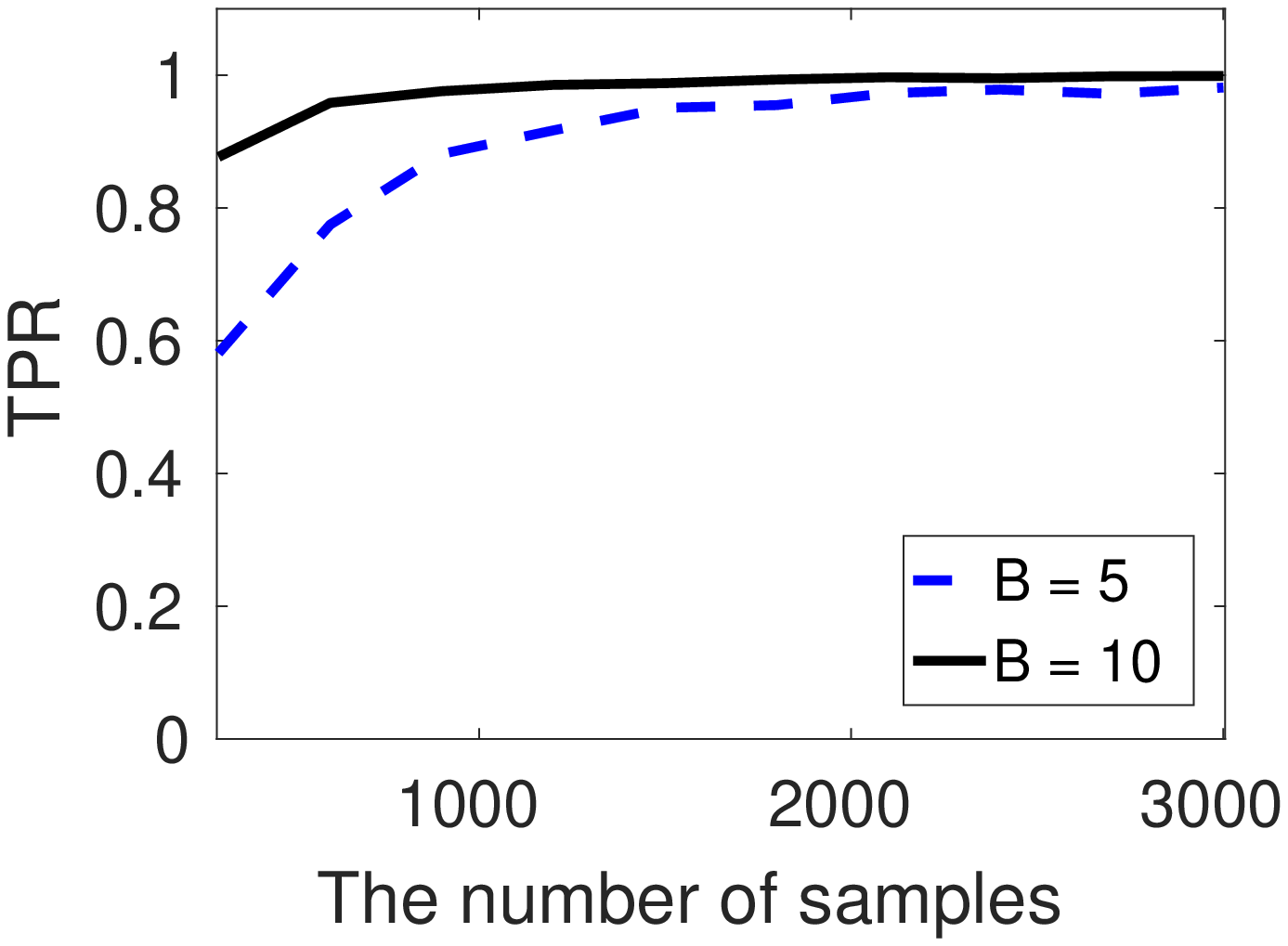}} \\ \vspace{-0.10cm}
(c) \texttt{hsicInf}  (TPR).
\end{minipage}
\begin{minipage}[t]{0.245\linewidth}
\centering
  {\includegraphics[width=0.99\textwidth]{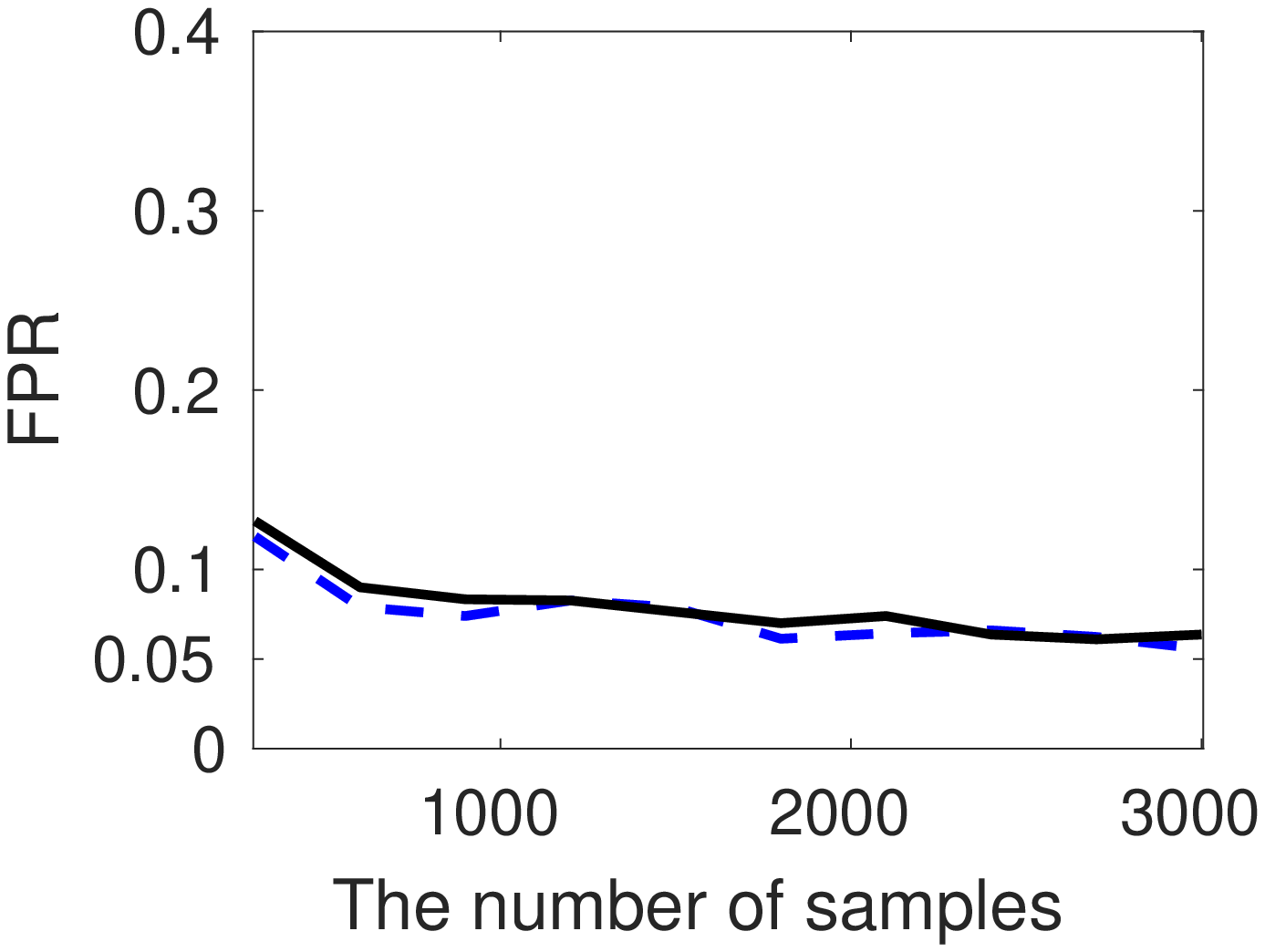}} \\ \vspace{-0.10cm}
(d) \texttt{hsicInf}  (FPR).
\end{minipage}
 \caption{The results for the multi-variate regression dataset. (a)(b): TPRs and FPRs of \texttt{hsicInf} ($B=10$). (c)(d): TPRs and FPRs  of \texttt{hsicInf} with different block size $B$.}
    \label{fig:synth3}
\end{center}
\vspace{-.2in}
\end{figure*}

\begin{figure*}[t!]
\begin{center}
\begin{minipage}[t]{0.245\linewidth}
\centering
  {\includegraphics[width=0.99\textwidth]{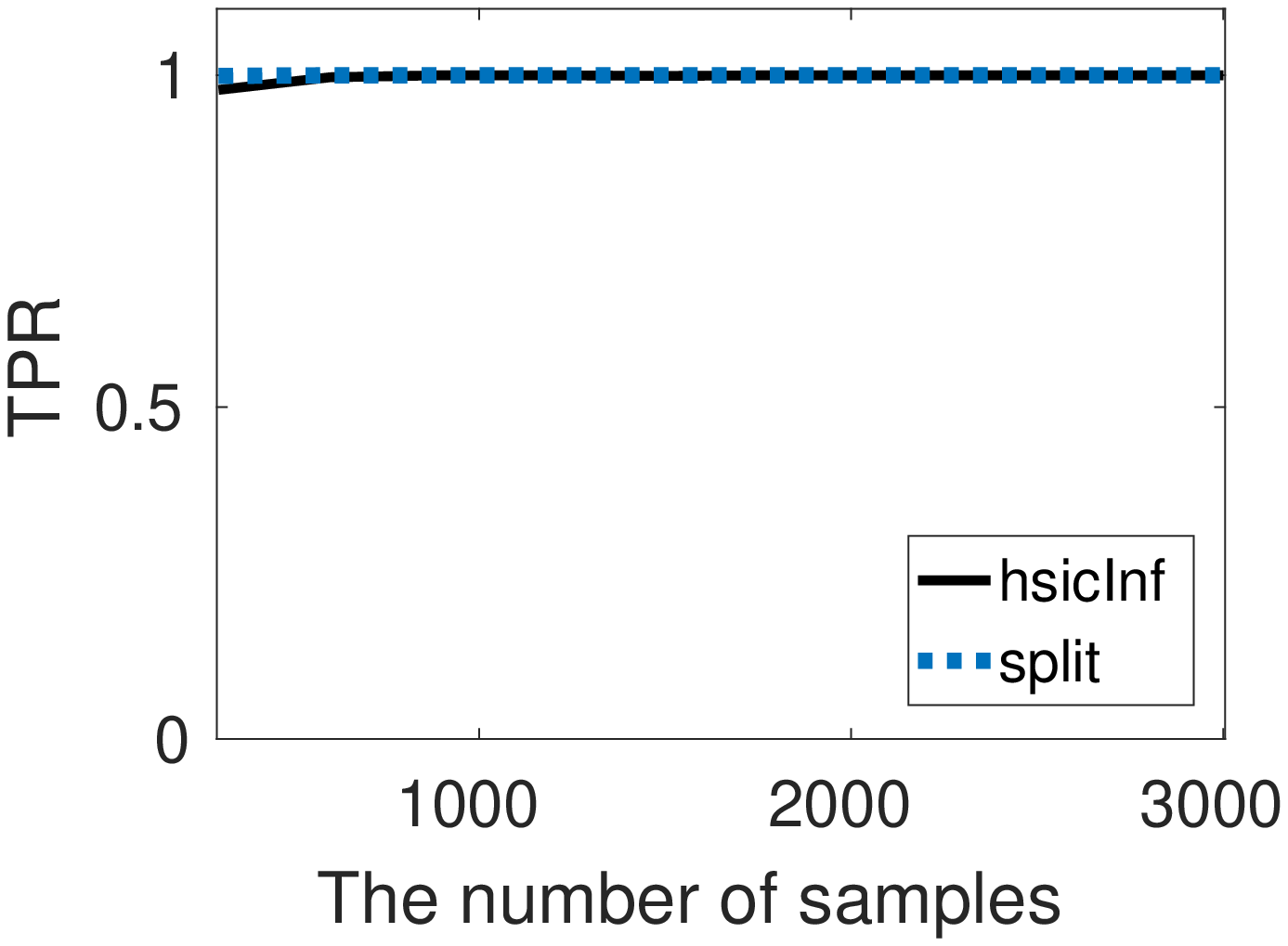}} \\ \vspace{-0.10cm}
(a) Multi-class (TPR).
\end{minipage}
\begin{minipage}[t]{0.245\linewidth}
\centering
  {\includegraphics[width=0.99\textwidth]{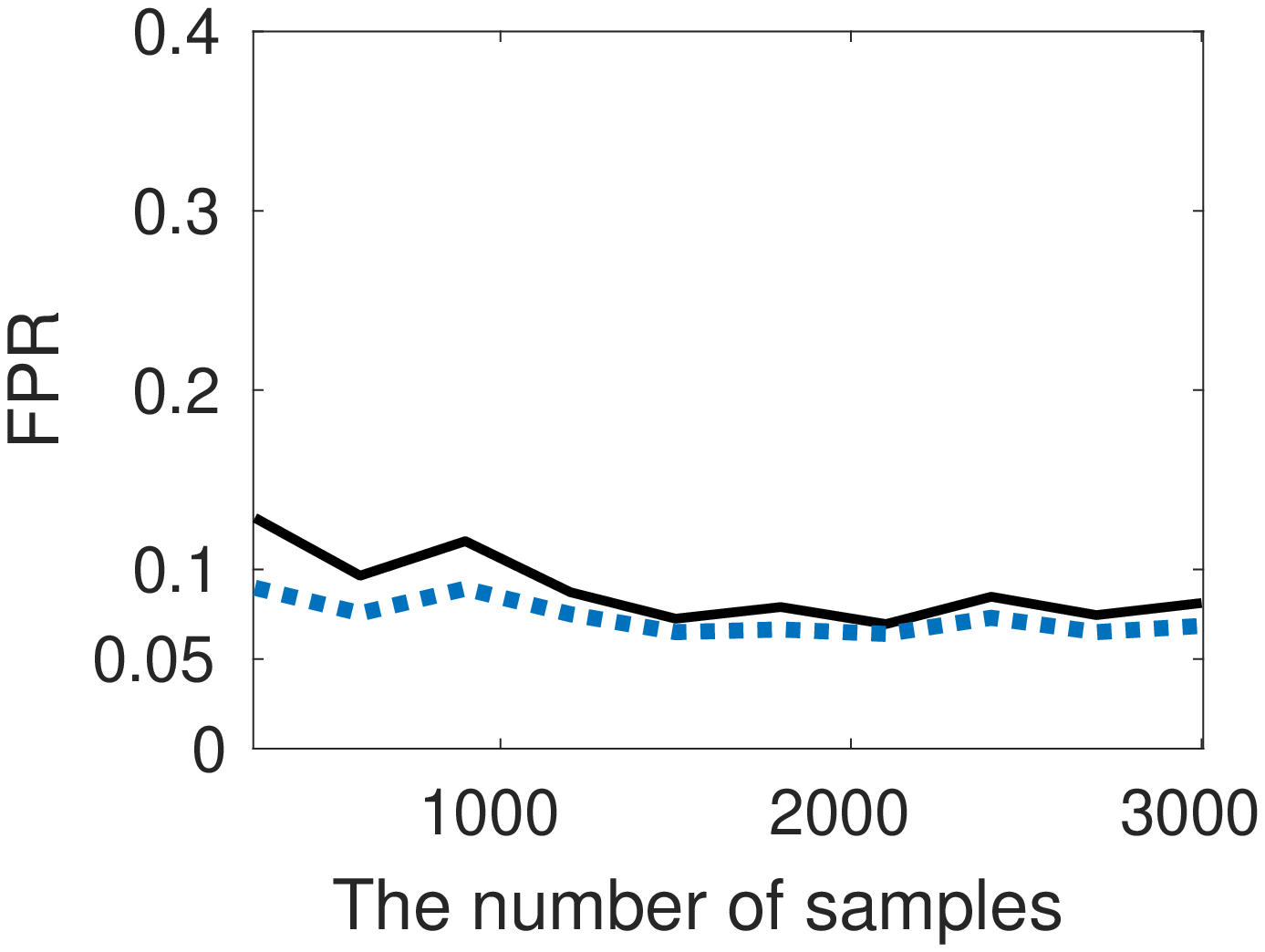}} \\ \vspace{-0.10cm}
(b) Multi-class (FPR).
\end{minipage}
\begin{minipage}[t]{0.245\linewidth}
\centering
  {\includegraphics[width=0.99\textwidth]{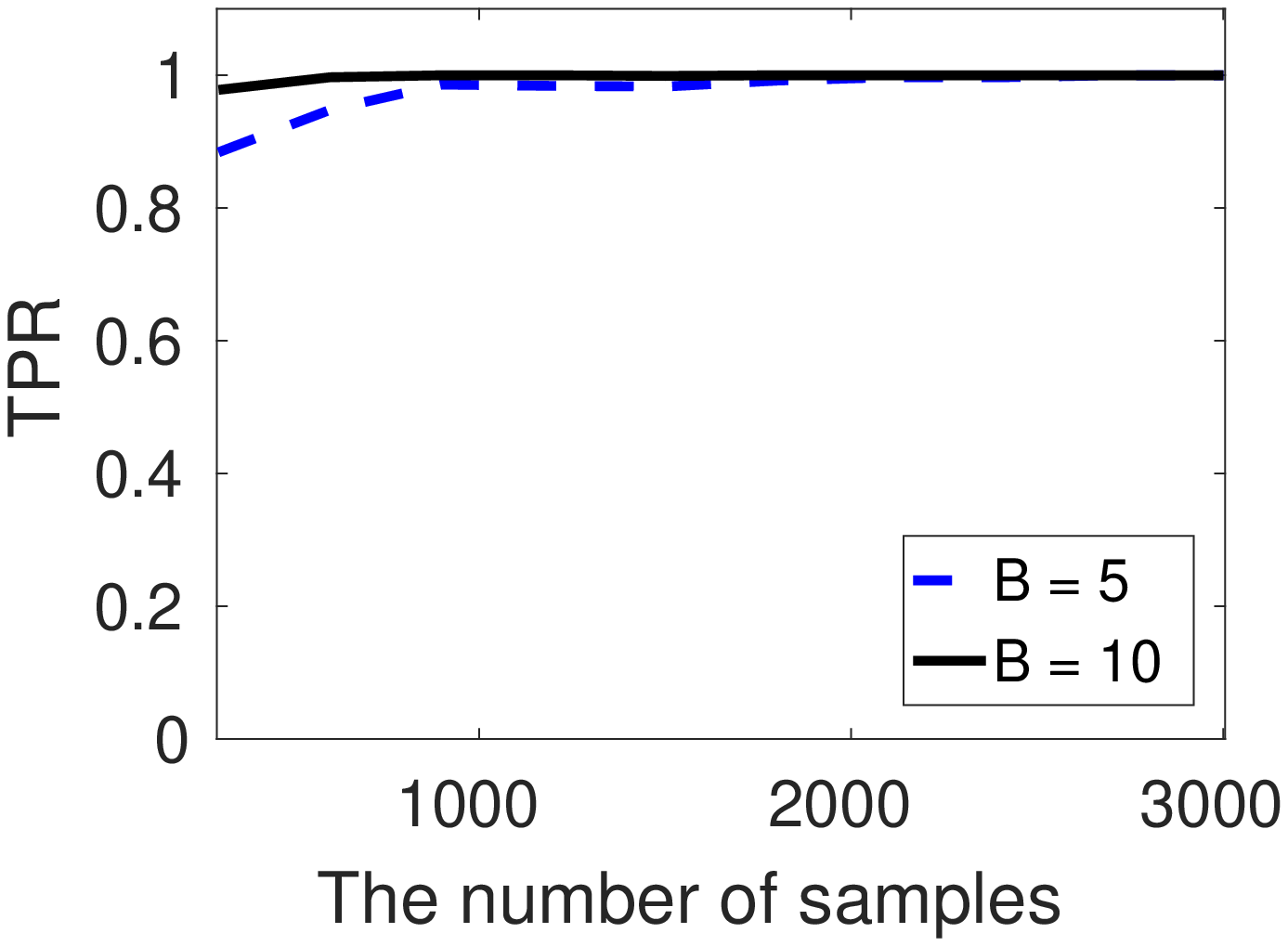}} \\ \vspace{-0.10cm}
(c) \texttt{hsicInf} (TPR).
\end{minipage}
\begin{minipage}[t]{0.245\linewidth}
\centering
  {\includegraphics[width=0.99\textwidth]{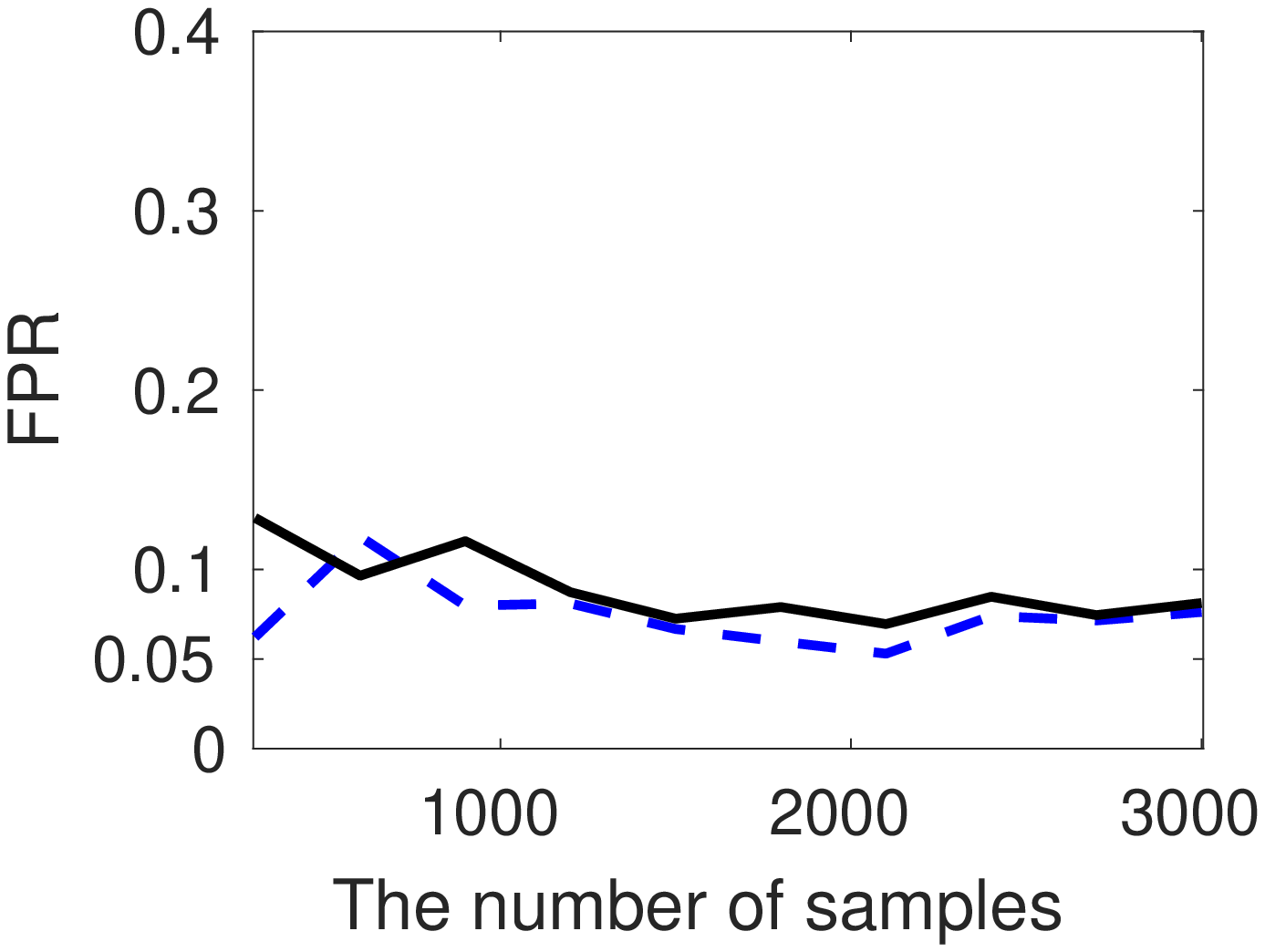}} \\ \vspace{-0.10cm}
(d) \texttt{hsicInf} (FPR).
\end{minipage}
 \caption{The results for the multi-class classification dataset.(a)(b): TPRs and FPRs of \texttt{hsicInf} ($B=10$). (c)(d): TPRs and FPRs  of \texttt{hsicInf} with different block size $B$.}
    \label{fig:classification}
\end{center}
\vspace{-.15in}
\end{figure*}

\subsection{True positive rate comparison}
Next, we compared the TPRs of \texttt{hsicInf}, \texttt{split}, and \texttt{larInf}.

\subsubsection{Synthetic Data (Regression)}
First, we evaluated whether the proposed method can find a  set of \emph{statistically significant} features from uni-variate linear/non-linear regression problems.

For this experiment, we first generated the input matrix $\boldX = [\boldx_1, \ldots, \boldx_n] \in \mathbbR^{20 \times n}$  where $\boldx \sim N(\boldzero, \bar{\boldSigma})$, $ [\bar{\boldSigma}]_{ij} = 0.95\delta_{ij} + 0.05, i,j \in \{1, 2, 3, 4, 5\}, [\bar{\boldSigma}]_{ii} = \delta_{ij},  i,j \in \{6,\ldots,20\}$, $\delta_{ij} = 1$ if $i = j$ and 0 otherwise, and $n = \{300, 600, \ldots, 3000\}$.

Then, we generated the corresponding output variable as
\begin{itemize}
\item {\bf Linear:} $Y = \sum_{i = 1}^5 X_i  + 0.1E$,
\item {\bf Additive Non-linear:} $Y = \sum_{i = 1}^5 X_i^2 + 0.1E$,
\item {\bf Non-additive Non-linear:} $Y = X_{1}\exp(X_{2})X_3\exp(X_4)X_5+ 0.1E$,
\end{itemize}
where $E \sim N(0,1)$ is a random variable. 

Figure~\ref{fig:synth1} (a)-(c) show TPRs of all methods. As we expected, the proposed \texttt{hsicInf} has higher TPRs compared to the data-splitting method \texttt{split}, since the proposed method can use larger number of samples than \texttt{split} for selecting features. Figure~\ref{fig:synth1} (d)-(f) show FPRs of all methods. Here, all the HSIC based approaches have larger FPRs than the significance level when the number of samples are small. This is due to the violation of the Gaussian assumption. \texttt{hsicInf} with small number of samples is an important future work.

The linear method \texttt{larInf} can select features for only linear setups, and it fails for non-linear counterparts. In contrast, the proposed algorithm can successfully detect statistically significant features for all setups.

 Figures~\ref{fig:synth1_b_rank} shows that TPRs and FPRs of \texttt{hsicInf}  with different block size $B$, respectively. These results indicate that the larger $B$ is usually preferable for having high detection power.

\begin{table*}
\caption{The $p$-values computed by \texttt{hsicInf} from the Turkiye Student Evaluation dataset.  \label{tab:hsicinf_turkiye}}
\begin{center}
\begin{tabular}{|l|c|}
\hline
Feature description & $p$-value \\ \hline \hline
{\bf The Instructor treated all students in a right and objective manner.} & $>0.001$\\
{\bf The Instructor arrived on time for classes.} & 0.033\\
{\bf The Instructor's knowledge was relevant and up to date.} & 0.018\\
{\bf The Instructor was open and respectful of the views of students about the course.} & 0.042\\
{\bf The Instructor demonstrated a positive approach to students. } & 0.033\\
The Instructor has a smooth and easy to follow delivery/speech. & 0.186\\
{\bf The Instructor encouraged participation in the course.} & 0.037\\
The Instructor's evaluation system  effectively measured the course objectives. & 0.176\\
The course aims and objectives were clearly stated at the beginning of the period. & 0.452 \\
The Instructor explained the course and was eager to be helpful to students. & 0.463 \\
\hline
\end{tabular}
\end{center}
\vspace{-.15in}
\end{table*}

\subsubsection{Synthetic Data (Multi-variate regression)}
We evaluated the proposed algorithm for multi-variate non-linear regression problems. Since  existing PSI methods cannot be used for multi-variate regression problems, we only reported the HSIC based approaches.

We used the zero-mean multi-variate Gaussinan input matrix with $ [\bar{\boldSigma}]_{ij} = 0.95\delta_{ij} + 0.05, i,j \in \{1, 2, 3, 4\}, [\bar{\boldSigma}]_{ii} = \delta_{ij},  i,j \in \{5,\ldots,20\}$. For the output variable,  we generated the three dimensional output variables as
\begin{align*}
Y_1 &= X_1 + 2X_{2} + 0.1E, \\
Y_2 &= 2X_1 + X_{2}^2  + 0.1E, \\ 
Y_3 &= X_{3} \exp(2X_{4}) + 0.1E.
\end{align*}
In this experiment, we have the training set $\{(\boldx_i,\boldy_i)\}_{i = 1}^n$ where $\boldx \in \mathbbR^{20}$ and $\boldy \in \mathbbR^3$. 


Figure~\ref{fig:synth3} shows TPRs and FPRs for the multi-variate regression case. As can be seen, both proposed methods can select statistically significant features.

\subsubsection{Synthetic Data (Multi-class Classification)}
We applied our proposed algorithm to a \emph{non-linear} three-class classification problem. Again, there is no existing multi-class PSI algorithm, and thus, we here simply reported the performance of the HSIC based algorithms.

In this experiment, we generated a three-class classification dataset as 
\begin{align*}
p(\boldx^{(1,2)} | y = 1) & = N\left(\left[
\begin{array}{c}
-3\\
0\\
\end{array}
\right],
\left[\begin{array}{cc}
1&0\\
0&1\\
\end{array}
\right]\right)\\
p(\boldx^{(1,2)} | y = 2) & =N\left(\left[
\begin{array}{c}
3\\
0\\
\end{array}
\right],
\left[\begin{array}{cc}
1&0\\
0&1\\
\end{array}
\right]\right)\\
p(\boldx^{(1,2)} | y = 3) & =0.5 N\left(\left[
\begin{array}{c}
0\\
3\\
\end{array}
\right],
\left[\begin{array}{cc}
1&0\\
0&2.25\\
\end{array}
\right]\right) \\
&\phantom{=}+ 0.5 N\left(\left[
\begin{array}{c}
0\\
-3\\
\end{array}
\right],
\left[\begin{array}{cc}
1&0\\
0&2.25\\
\end{array}
\right]\right).
\end{align*}
Then, we generated the final feature $\boldx = [(\boldx^{(1,2)})^\top~ \widetilde{\boldx}^\top]^\top$ where  $\widetilde{\boldx} \in \mathbbR^{18}$ and $\widetilde{\boldx} \sim N(\boldzero, \boldI)$.
 
 Figure~\ref{fig:classification} shows the TPR and FPR for the three-class classification problem, and the \texttt{hsicInf} algorithm can perfectly detect the important features.
 
\subsection{Real-world data}
Finally, we evaluated our proposed method on the Turkiye Student Evaluation Data Set \citep{GunduzFokoue:2013}. 

This dataset consists of 5820 samples with 28 features\footnote{\url{http://archive.ics.uci.edu/ml/datasets/Turkiye+Student+Evaluation#}}. We used the "Level of difficulty of the course as perceived by the student ($\{1,2,\ldots,5\}$)" as the output variable, and selected features that are significantly affected to the difficulty of the course. In this experiment, we set the number of selected features $k = 10$, the block size $B = 10$, and used the Gaussian kernel for output. Note that, since the output variable takes integers and are non-Gaussian, \texttt{larInf} cannot be used for this data.

Table~\ref{tab:hsicinf_turkiye} shows the selected features by \texttt{hsicInf}. As can be seen, the difficulties of class are highly related to the attitude of teachers and teacher's support to students, and this result is reasonable.

\section{Conclusion}
In this paper, we proposed a novel post selection inference (PSI) algorithm \texttt{hsicInf}. The key advantage of the proposed method is that it can select a \emph{statistically significant} features from non-linear and/or multi-variate data and have high detection power. To the best of our knowledge, this is the first work  to address the PSI algorithm for both nonlinear and multi-variate regression problems. Through several experiments, we showed that the proposed method outperformed a state-of-the-art \emph{linear} PSI algorithm.

{
\bibliography{main}
\bibliographystyle{plainnat}
}

\end{document}

%% file: mathdef.tex
\newtheorem{defi}{Definition}

\newtheorem{theo}[defi]{Theorem}

\newcommand{\proofend}{\hfill$\Box$\vspace{2mm}}

\newcommand{\mathbbE}{\mathbb{E}}

\newcommand{\mathbbR}{\mathbb{R}}

\newcommand{\boldzero}{{\boldsymbol{0}}}
\newcommand{\boldone}{{\boldsymbol{1}}}

\newcommand{\boldA}{{\boldsymbol{A}}}

\newcommand{\boldI}{{\boldsymbol{I}}}

\newcommand{\boldK}{{\boldsymbol{K}}}
\newcommand{\boldL}{{\boldsymbol{L}}}

\newcommand{\boldX}{{\boldsymbol{X}}}

\newcommand{\boldb}{{\boldsymbol{b}}}
\newcommand{\boldc}{{\boldsymbol{c}}}

\newcommand{\bolde}{{\boldsymbol{e}}}

\newcommand{\boldx}{{\boldsymbol{x}}}
\newcommand{\boldy}{{\boldsymbol{y}}}
\newcommand{\boldz}{{\boldsymbol{z}}}

\newcommand{\boldeta}{{\boldsymbol{\eta}}}

\newcommand{\boldmu}{{\boldsymbol{\mu}}}

\newcommand{\boldSigma}{{\boldsymbol{\Sigma}}}

\newcommand{\calD}{{\mathcal{D}}}

\newcommand{\calS}{{\mathcal{S}}}

\newcommand{\calZ}{{\mathcal{Z}}}

\newcommand{\boldcalZ}{{\bm \calZ}}

\newcommand{\ntr}{n}

